\documentclass[10pt,twocolumn,letterpaper]{article}

\usepackage{wacv}              %

\usepackage{graphicx}
\usepackage{amsmath}
\usepackage{amssymb}
\usepackage{booktabs}

\usepackage{tabularray}
\usepackage{tabularx, booktabs, relsize} %

\usepackage[accsupp]{axessibility} %

\usepackage[pagebackref,breaklinks,colorlinks]{hyperref}

\usepackage[capitalize]{cleveref}
\crefname{section}{Sec.}{Secs.}
\Crefname{section}{Section}{Sections}
\Crefname{table}{Table}{Tables}
\crefname{table}{Tab.}{Tabs.}

\def\wacvPaperID{246} %
\def\confName{WACV}
\def\confYear{2025}

\newif\ifarxiv

\arxivfalse

\ifarxiv
    \newcommand{\arxiv}[1]{#1}
    \newcommand{\nonarxiv}[1]{}
\else
    \newcommand{\arxiv}[1]{}
    \newcommand{\nonarxiv}[1]{#1}
\fi

\begin{document}

\title{Detecting Origin Attribution for Text-to-Image Diffusion Models}

\author{%
  Katherine Xu\\
  University of Pennsylvania\\
  \and
  Lingzhi Zhang\\
  Adobe Inc.\\
  \and
  Jianbo Shi\\
  University of Pennsylvania\\
}

\maketitle

\begin{abstract}
    \vspace{-10pt}
    Modern text-to-image (T2I) diffusion models can generate images with remarkable realism and creativity. These advancements have sparked research in fake image detection and attribution, yet prior studies have not fully explored the practical and scientific dimensions of this task. In addition to attributing images to 12 state-of-the-art T2I generators, we provide extensive analyses on what inference stage hyperparameters and image modifications are discernible. Our experiments reveal that initialization seeds are highly detectable, along with other subtle variations in the image generation process to some extent. We further investigate what visual traces are leveraged in image attribution by perturbing high-frequency details and employing mid-level representations of image style and structure. Notably, altering high-frequency information causes only slight reductions in accuracy, and training an attributor on style representations outperforms training on RGB images. Our analyses underscore that fake images are detectable and attributable at various levels of visual granularity.
    \vspace{-15pt}
\end{abstract}

\section{Introduction}
\label{sec:intro}
\vspace{-5pt}

In recent years, the emergence of advanced text-to-image (T2I) diffusion models \cite{rombach2022highresolution_stable_diffusion, podell2023sdxl, sauer2023adversarial, luo2023latent_consistency_model_lcm, pernias2023wuerstchen, razzhigaev2023kandinsky, ramesh2022hierarchical_dalle, betker2023improving} has markedly transformed the landscape of image generation. These advancements enable the creation of highly realistic and imaginative visual content directly from textual descriptions, heralding new possibilities for creative expression and practical applications. However, this progress also introduces significant challenges in discerning real images from AI-generated images and accurately identifying their origins. Addressing these challenges is vital for preserving the integrity of visual content across digital platforms.

Previous studies \cite{wang2020cnngenerated, bi2023detecting, sinitsa2023deep, chai2020makes, ju2022fusing, amoroso2023parents, zhong2024patchcraft} have focused on differentiating AI-generated images from real ones, with some research attributing images to their source generators, notably in GAN variants \cite{yu2019attributing, marra2019incremental, girish2021discovery, bui2022repmix} and diffusion models \cite{corvi2022detection, sha2023defake, guarnera2023level}. Yet, these investigations have largely been conducted using generative models that may not reflect the latest advancements in the field. Moreover, these studies have not fully explored the broader, practical, and scientific dimensions of these tasks, which we aim to further examine. 

As a first step, our work unifies ``real vs. fake" classification and image attribution into a single task by simply treating real images as an additional category. We expand the analysis to include a comprehensive range of state-of-the-art T2I diffusion models, as of July 2024. This includes Stable Diffusion (SD) 1.5 \cite{rombach2022highresolution_stable_diffusion}, SD 2.0 \cite{rombach2022highresolution_stable_diffusion}, SDXL \cite{podell2023sdxl}, SDXL Turbo \cite{sauer2023adversarial}, Latent Consistency Model (LCM) \cite{luo2023latent_consistency_model_lcm}, Stable Cascade \cite{pernias2023wuerstchen}, Kandinsky 2.1 \cite{razzhigaev2023kandinsky}, DALL-E 2 \cite{ramesh2022hierarchical_dalle}, DALL-E 3 \cite{betker2023improving}, along with Midjourney versions 5.2 and 6 \cite{Midjourney}. To encompass a variety of visual concepts, we utilize 5,000 captions from MS-COCO \cite{lin2014microsoft} for natural scenes and employ GPT-4 \cite{achiam2023gpt} to generate another 5,000 creative and surreal prompts. For each prompt, we generate multiple images from each model, amassing a dataset of nearly half a million AI-generated images to train our image attributor, where the details are discussed in Sec. \ref{sec:dataset}. Regarding performance, our top-performing attributor reaches an accuracy exceeding 90\%, significantly surpassing the baseline random chance of merely 7.69\%, as detailed in Sec. \ref{sec:training_image_attributors}.

Moving beyond previous research that focused on attributing images to their originating generators, our study probes further into whether nuanced changes in hyperparameters during the inference phase of the same T2I diffusion model can be identified. We examine hyperparameters including model checkpoints at different training iterations, scheduler types, the number of sampling steps, and initialization seeds. A significant finding from our experiments is the ability to distinguish between initialization seeds with 98\%+ accuracy, employing ten unique seeds for image generation within a consistent generator framework. While the accuracy in identifying other hyperparameters doesn't reach the exceptional levels observed with initialization seeds, they all notably exceed random chance. This suggests that even subtle variations in the generation process can indeed be discerned to some extent. More details are in Sec. \ref{sec:Analyzing_the_Detectability_of_Hyperparameter_Variations}.

In a workflow involving AI-generated images, users often enhance these images further by importing them into additional software or models for regional editing via SDXL Inpainting \cite{podell2023sdxl} or Photoshop Generative Fill (Ps GenFill) \cite{Photoshop}, or employing tools like Magnific AI \cite{MagnificAI} for texture enhancement at higher resolutions. This raises an essential question: Can we still trace these post-edited images back to their original generators, and to what extent is this feasible? In Sec. \ref{sec:Assessing_Detectability_Following_Post-Editing_Enhancements}, we mimic user-driven regional editing using SDXL Inpainting and Ps GenFill, alongside utilizing Magnific AI on a set of test images. Our discussion thoroughly examines and provides insights into how these post-editing interventions impact the image attribution performance.

Lastly, while prior research has showed notable success in discerning ``real vs. fake" images and accurately attributing them to their origins, the exact nature of detectable traces recognized by classifiers and their locations within images remain elusive. In Sec. \ref{sec:Detecting_Image_Attribution_Beyond_RGB}, we study this scientific question by introducing perturbations in the high-frequency domain and converting images into mid-level representations, such as depth maps and Canny edges, to assess their impact on image attribution accuracy. This strategy aims to unearth detectable traces across levels of visual granularity, enriching our understanding of how classifiers identify and attribute images. Notably, our investigations reveal that training the image attributor using style representations---specifically, the Gram matrix---enhances accuracy beyond what is achievable with attributors trained on original RGB images. Moreover, adding perturbations to high-frequency signals within images results in only minor performance decreases in the attributors. When these attributors are trained on mid-level representations, they maintain commendable accuracy levels that surpass random chance. This observation suggests that detectable traces extend beyond the high-frequency domain, encompassing mid-level aspects of texture, structure, and potentially the layout of images.

Overall, our key contributions are as follows:
\vspace{-8pt}
\begin{itemize}
    \renewcommand{\labelitemi}{\LARGE$\cdot$}
    \itemsep-0.2em
    
    \item Developed an extensive dataset of nearly half a million AI-generated images from cutting-edge T2I models with a variety of natural and surreal prompts.
    
    \item Achieved over 90\% accuracy in training an image attributor across 12 contemporary T2I generators and real images, significantly outperforming random chance for the 13-way classification task.
    
    \item Pioneered the exploration of detectability regarding minor hyperparameter modifications during the inference stage of T2I diffusion models.
    
    \item Innovatively replicated user editing workflows on AI-generated images using various tools, thoroughly evaluating their effect on attribution accuracy.
    
    \item Introduced a novel approach for analyzing detectable traces within images through high-frequency perturbations and conversion to diverse mid-level representations, yielding significant insights.
    
\end{itemize}

\vspace{-14pt}
\section{Related Work}
\label{sec:related}
\vspace{-6pt}

\textbf{Classifying Fake vs. Real Images.}
The rise of sophisticated image generators facilitates creating highly realistic images with diverse styles, which has spurred research in detecting synthetic images from real images. Wang \etal \cite{wang2020cnngenerated} introduced a CNN that identifies images from GANs \cite{goodfellow2014generative} and low-level vision models \cite{chen2018learning, dai2019second}. They showed that training diversity is crucial for fake image detectors to achieve good generalization. Additionally, Yu \etal \cite{yu2019attributing} discovered that different GAN architectures, training sets, and initialization seeds lead to distinct image fingerprints.

Various approaches detect synthetic images using visible \cite{exploit_visual} and invisible artifacts that can lie in the spatial or frequency domain \cite{gragnaniello2021gan}.
Spatial domain methods often estimate these digital fingerprints using deep learning methods \cite{bi2023detecting, sinitsa2023deep} or by averaging their noise residuals \cite{marra2018gans}. These detection methods may use local image patches \cite{chai2020makes}, combine local and global image features \cite{ju2022fusing}, or use gradients extracted by a pretrained CNN \cite{tan2023learning}.
Moreover, style and texture information have been utilized for fake image detection \cite{liu2020global, zhong2024patchcraft, amoroso2023parents}. Amoroso \etal revealed that real and fake images are more easily separable using style features rather than semantics \cite{amoroso2023parents}, and Zhong \etal \cite{zhong2024patchcraft} found it more challenging for generative models to synthesize rich texture regions.
Another line of work suggests that GAN-generated images can be detected by studying artifacts in the frequency domain \cite{corvi2023intriguing, ricker2024detection, bi2023detecting, think_twice, frequency_aware, marra2018gans, frank2020leveraging, bammey2023synthbuster, durall2020watch, dzanic2020fourier}.

Recent works have aimed at identifying images generated by diffusion models \cite{sha2023defake, wang2023dire, ha2024organic, bird2023cifake, zhang2023diffusion, zhu2023gendet, epstein2023online}. Wang \etal \cite{wang2023dire} discovered that features of diffusion-generated images are more easily reconstructed by pretrained diffusion models than real images, and Cozzolino \etal \cite{cozzolino2023raising} observed that images from diffusion models have spectral peaks that distinguish them from real images. Furthermore, the idea of learning classifiers that leverage both visual and language features to supplement low-level features has gained interest \cite{wu2023generalizable, cozzolino2023raising, ojha2023universal}. Ojha \etal \cite{ojha2023universal} found that using the feature space of CLIP \cite{radford2021learning} improves generalization ability for detecting fake images from GANs and diffusion models.

\textbf{Detecting Fake Image Attribution.}
In addition to recognizing synthetic images, some works strive to identify the source of generated images. Yu \etal \cite{yu2019attributing} discovered that different GAN architectures, training sets, and initialization seeds can lead to fingerprint features for attribution. RepMix \cite{bui2022repmix} traces GAN images to their generators while being invariant to semantic content and image perturbations. Girish \etal \cite{girish2021discovery} and Marra \etal \cite{marra2019incremental} developed algorithms for online detection and attribution of GAN images. Recent work has explored fake image detection and attribution from diffusion models and T2I models \cite{corvi2022detection, laszkiewicz2023single, sha2023defake, wang2024did, liu2024model}.
Guarnera \etal \cite{guarnera2023level} proposed a method to classify images as real or fake, GAN or diffusion-generated, and the specific generator. Guo \etal \cite{guo2023hierarchical} takes a similar approach but also determines whether the image was only partially edited.

\begin{figure*}[!h]
 \vspace{-24 pt}
    \centering
    \includegraphics[trim=0in 4.5in 0in 0in, clip,width=\textwidth]{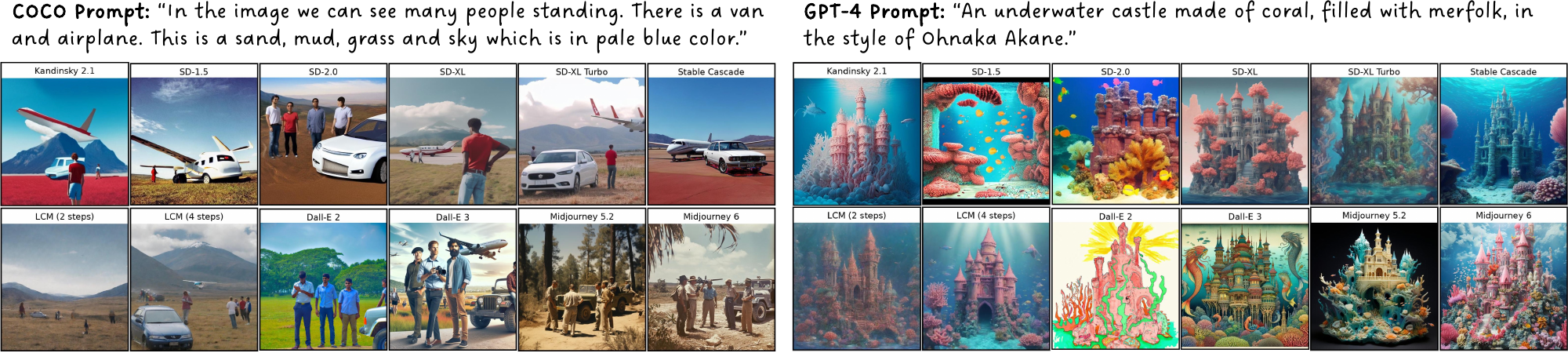}
    \vspace{-20 pt}
    \caption{A depiction of images generated for our dataset, showcasing two types of prompts: MS-COCO \cite{lin2014microsoft} derived captions on the left, and creative prompts generated by GPT-4 on the right. For both categories, images were produced using 12 different T2I generators. }
    \label{fig:gen_from_models_prompts}
    \vspace{-8 pt}
\end{figure*}

\begin{figure*}[!h]
 \vspace{-5 pt}
    \centering
    \includegraphics[trim=0in 4.45in 0in 0in, clip,width=\textwidth]{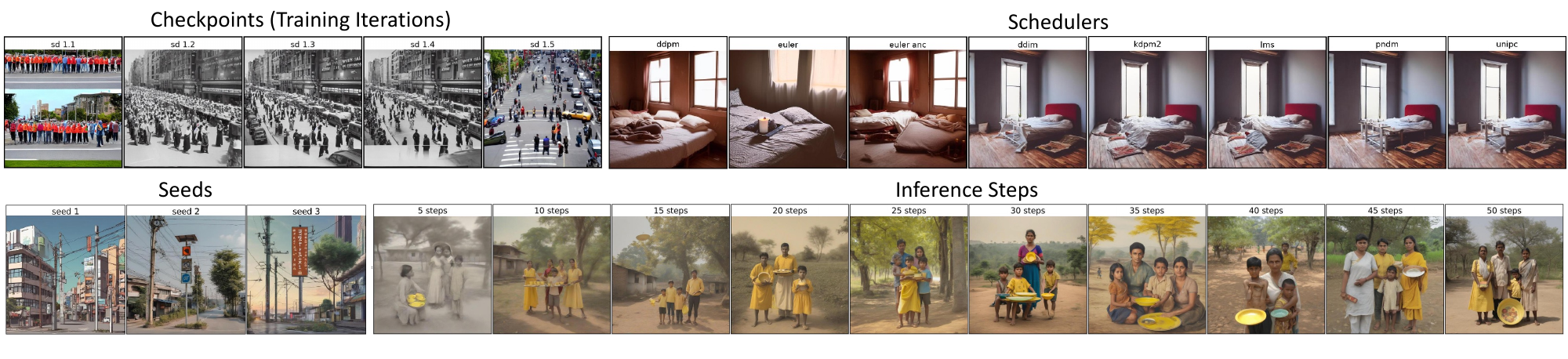}
    \vspace{-26 pt}
    \caption{An illustration showcasing the diversity in generated images influenced by varying hyperparameters: different model checkpoints (within the same architecture), diverse scheduling algorithms, varied initialization seeds, and a range of inference steps. }
    \label{fig:gen_from_hyperparameter_variations}
    \vspace{-18 pt}
\end{figure*}

\textbf{Analyzing Images Generated by Diffusion Models.}
Prior works have studied scene knowledge within pretrained diffusion models \cite{zhan2023does_sd, chen2023surface, du2023generative} and examined the geometry of diffusion-generated images \cite{sarkar2023shadows}. Du \etal \cite{du2023generative} discovered that generative models contain rich information about scene intrinsics, and they train a low-rank adapter \cite{hu2021lora} to produce surface normals, shading, albedo, and depth. Sarkar \etal \cite{sarkar2023shadows} revealed that synthetic images can be differentiated from real ones by analyzing their geometric properties.

In contrast, our work further examines the specific inference stage hyperparameters, image modifications, and levels of visual granularity discernible by an image attributor.

\vspace{-6 pt}
\section{Dataset Generation}
\label{sec:dataset}
\vspace{-6 pt}

In this work, we detect image attributions for modern text-to-image (T2I) models, while also investigating the extent to which traces can be detected across different generators and over inference stage controls. To achieve this, we first generate images using a variety of T2I models, employing a wide range of text prompts to ensure diversity. Next, we maintain a consistent generator while adjusting inference time hyperparameters, such as the number of inference steps, scheduler, model checkpoints, and random seeds.

\vspace{-4 pt}
\subsection{Images from Diverse Generators and Prompts}
\label{sec: images_from_diverse_generators_and_prompts}
\vspace{-4 pt}

As of July 2024, we have employed these state-of-the-art, open-source T2I models for image generation: SD 1.5 \cite{rombach2022highresolution_stable_diffusion}, SD 2.0 \cite{rombach2022highresolution_stable_diffusion}, SDXL \cite{podell2023sdxl}, SDXL Turbo \cite{sauer2023adversarial}, Latent Consistency Model (LCM) \cite{luo2023latent_consistency_model_lcm}, Stable Cascade \cite{pernias2023wuerstchen}, Kandinsky 2 \cite{razzhigaev2023kandinsky}, DALL-E 2 \cite{ramesh2022hierarchical_dalle}, DALL-E 3 \cite{betker2023improving}, and Midjourney versions 5.2 and 6 \cite{Midjourney}. To generate images, we use the OpenAI API for DALL-E 2 and 3, an automation bot for Midjourney 5.2 and 6, and the Hugging Face diffusers GitHub repository \cite{von-platen-etal-2022-diffusers} for the remaining models.

To gather a broad spectrum of text prompts, we include both descriptions of natural scenes and imaginative, surreal prompts. This diversity is achieved by leveraging around 5,000 captions from the MS-COCO dataset \cite{lin2014microsoft}, complemented by around 5,000 prompts generated by GPT-4 \cite{achiam2023gpt}. The GPT-4 generated prompts stem from a wide-ranging collection of popular user prompts found online, details of which are provided in the supplemental materials. Utilizing this comprehensive set of prompts, we generate images across all the text-to-image (T2I) models referenced, as depicted in Fig. \ref{fig:gen_from_models_prompts}. For each prompt, we generated one image for DALL-E 2 and 3 (due to cost considerations), four images for Midjourney, and five images for the other models, culminating in a dataset exceeding 450K generated images. It's important to note that not all images were used during training; the specifics are in the supplemental materials.

\begin{figure*}[!h]
 \vspace{-24 pt}
    \centering
    \includegraphics[trim=0in 3.55in 0in 0in, clip,width=\textwidth]{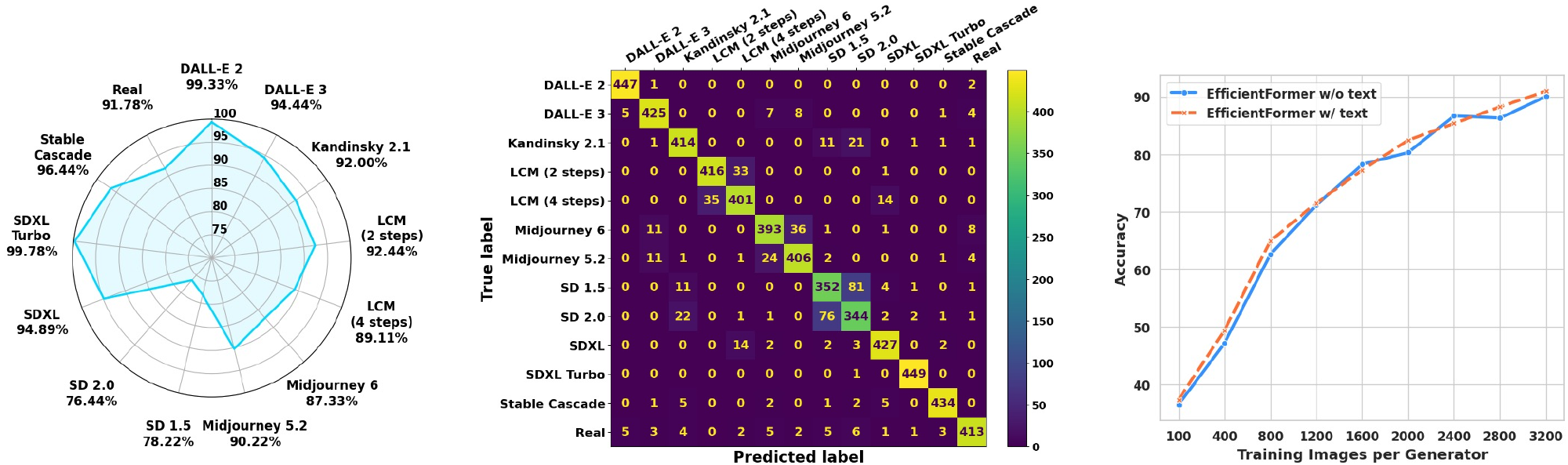}
    \vspace{-18 pt}
    \caption{ \textbf{Left/Middle:} Accuracy and confusion matrix of EfficientFormer trained with text prompts, which achieved the highest accuracy in Table \ref{tab:image_attributor_metric}. \textbf{Right:} Accuracy of EfficientFormer as we vary the number of training images.}
    \label{fig:image_attributor_results}
    \vspace{-18 pt}
\end{figure*}

\vspace{-6 pt}
\subsection{Images from Varying Hyperparameters During Inference Stage}
\vspace{-5 pt}

In this research, we expand our focus beyond simply identifying the source generators based on their architectures, to a deeper analysis of the critical yet subtle choices made during the inference stage that have a profound effect on the generated outputs. Initially, we investigate the possibility of identifying specific model checkpoints within the same architecture, specifically Stable Diffusion (SD) \cite{rombach2022highresolution_stable_diffusion}, based on different training iterations. To facilitate this, we generated images using five versions of SD from 1.1 to 1.5. Despite sharing a common architecture, each version was trained for a distinct number of iterations. Next, we explore the impact of using different schedulers or samplers \cite{ho2020denoising, song2020denoising, karras2022elucidating, zhao2024unipc} during the inference phase for the same generator. We question whether the generated images can reveal which scheduler was employed. Furthermore, drawing inspiration from studies indicating that the use of different seeds in GAN-generated images can be detected \cite{yu2021artificial}, we seek to apply this concept to diffusion models to determine if the seed number is detectable in the resulting images. Finally, we conduct experiments with diffusion steps ranging from 5 to 50 in increments of 5 to investigate whether the number of sampling steps can leave detectable traces in the images. Selected samples of images generated under different hyperparameter adjustments are presented in Fig. \ref{fig:gen_from_hyperparameter_variations}.

\vspace{-8pt}
\section{Detecting Image Attribution in RGB}
\label{sec:method}
\vspace{-8pt}

In this section, we benchmark image attribution performance across 12 modern text-to-image generators, examining the impact of various architectures, training sizes, and cross-domain influences on task performance. We then analyze the detectability of traces for various hyperparameter adjustments during inference time. Finally, inspired by typical user workflows, we investigate whether AI-generated images can still be attributed to their original generators after being modified by distinct software or models.
\begin{figure*}[!h]
 \vspace{-24 pt}
    \centering
    \includegraphics[trim=0in 4.16in 0.7in 0in, clip,width=\textwidth]{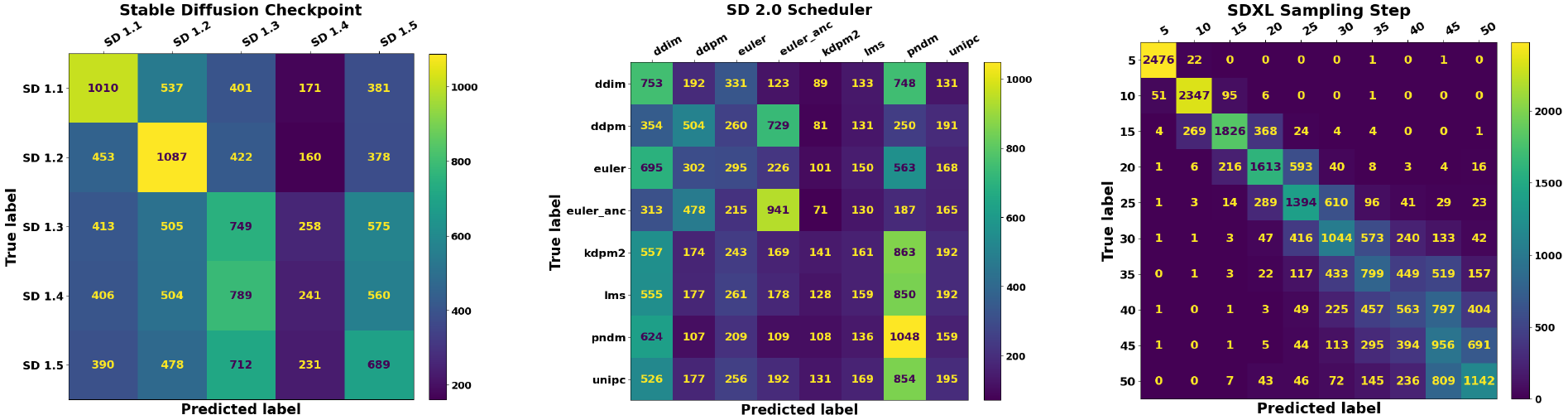}
    \vspace{-19 pt}
    \caption{Confusion matrices for hyperparameter variations, including Stable Diffusion version, scheduler, and number of inference steps. We observe that images generated with fewer SDXL sampling steps are more detectable, likely due to visible degradation in image quality.}
    \label{fig:confusion_hyperparam}
    \vspace{-14 pt}
\end{figure*}

\vspace{-4pt}
\subsection{Training Image Attributors}
\label{sec:training_image_attributors}
\vspace{-4pt}

\textbf{Problem Setup and Model Performance. } Prior research has demonstrated deep networks' ability to distinguish AI-generated images from real ones \cite{exploit_visual, marra2018gans, wang2020cnngenerated, chai2020makes, ju2022fusing, ojha2023universal, cozzolino2023raising} and identify their sources \cite{yu2019attributing, bui2022repmix, sha2023defake}. Our study builds on this foundation by merging the tasks of discerning ``AI-generated vs. Real Images" and attributing images to their sources into a single framework. This is achieved by including real images in our dataset and treating them as an additional `generator', enabling a more detailed analysis of AI-generated content. Concerning the architecture of the image attributor, which functions as an image classifier, previous studies \cite{ojha2023universal, cozzolino2023raising} have demonstrated that a straightforward linear probe or nearest neighbor search, when applied to a large pretrained model like CLIP \cite{radford2021learning}, can effectively differentiate AI-generated images from real ones. Inspired by these findings, we employ three network architectures to tackle the attribution task across 12 modern text-to-image (T2I) generators---such as SDXL Turbo \cite{sauer2023adversarial}, DALL-E 3 \cite{betker2023improving}, and Midjourney 6---plus a real image dataset. These architectures include an EfficientFormer \cite{li2022efficientformer} trained from scratch, a CLIP \cite{radford2021learning} backbone connected with a linear probe and MLP, and DINOv2 \cite{oquab2023dinov2} with a similar configuration. We also analyze the impact of incorporating text prompts as additional inputs similar to Sha \etal \cite{sha2023defake}, which we found to provide slight yet consistent improvements across all architectures, as shown in Tab. \ref{tab:image_attributor_metric}.

\begin{table}[h!]
\centering
{\smaller %
\begin{tabularx}{\columnwidth}{l| *{4}{>{\centering\arraybackslash}X|} >{\centering\arraybackslash}X}
\toprule
          & E.F. (scratch) & CLIP + LP  & CLIP + MLP & DINOv2 + LP & DINOv2 + MLP \\
\midrule
w/o text  & 90.03\%        & 70.15\%  & 73.09\%  & 67.68\%   & 71.33\%       \\
w/ text   & 90.96\%        & 71.44\%  & 74.19\%  & 69.44\%   & 73.08\%       \\
\bottomrule
\end{tabularx}
}
\vspace{-8 pt}
\caption{The 13-way classification accuracy of various architectures for image attribution learning performed across 12 generators and a corresponding set of real images, with each category containing an equal number of images. The probability of randomly guessing the correct source is $\frac{1}{13}$, which gives a $\textbf{7.69\%}$ accuracy. In this context, ``E.F." refers to EfficientFormer trained from scratch. The first and second rows in the results table indicate classifiers trained without and with text prompts, respectively.}
\label{tab:image_attributor_metric}
\vspace{-10 pt}
\end{table}

\textbf{Classifier Performance Across Generators. } To provide a more granular view of our analysis, we analyze the performance of each classifier, illustrating a detailed accuracy breakdown through a radar graph and a corresponding confusion matrix in Fig. \ref{fig:image_attributor_results}.
Our findings reveal a noticeable challenge in differentiating between generators from the same family, with notable pairs including ``SD 1.5 vs. SD 2.0," ``Midjourney 5.2 vs. Midjourney 6," and ``LCM (2 steps) vs. LCM (4 steps)."
While Midjourney's architecture remains undisclosed to the public, it is reasonable to infer that versions 5.2 and 6 likely share a similar underlying architecture from our analysis.
Interestingly, DALL-E 3 presents more confusion when compared to Midjourney versions 5.2 / 6, rather than with DALL-E 2. We attribute this to the significant architectural differences: DALL-E 2 incorporates pixel diffusion in its decoder stage, whereas DALL-E 3 employs multi-stage latent diffusion alongside a distinct one-step VAE decoder, similar to \cite{rombach2022highresolution_stable_diffusion}, leading to divergent generative characteristics.
Finally, we demonstrate that the accuracy of the attributor consistently improves with an increase in the number of training images, as shown on the right side of Fig. \ref{fig:image_attributor_results}. However, due to budget constraints, fully exploring the dataset expansion up to the saturation point is deferred to future research endeavors.

\textbf{Cross-domain Generalization. } As highlighted in Sec. \ref{sec: images_from_diverse_generators_and_prompts}, user prompts vary significantly, with some describing natural scenes and others depicting surreal concepts. This diversity led us to examine how a classifier, trained on images generated using MS-COCO captions, would perform when applied to images created from GPT-4's inventive prompts, and conversely. The results, presented in Tab. \ref{tab:cross_domain_classification}, show an evident decline in performance when the classifier is trained and tested across these differing domains. Since we keep the same generators and only change the style of prompts, this outcome underscores that learning image attribution uses the visible content in the generated images.

\begin{table}[h!]
\vspace{-8 pt}
\centering
{\smaller %
\begin{tabularx}{\columnwidth}{l| *{2}{>{\centering\arraybackslash}X|} >{\centering\arraybackslash}X}
\toprule
            & Train on MS-COCO & Train on GPT-4 & Train on Both \\
\midrule
Test on MS-COCO   & 89.04\%  & 71.07\%  & 85.78\%  \\
Test on GPT-4     & 69.24\%  & 79.35\%  & 81.06\%  \\
\bottomrule
\end{tabularx}
}
\vspace{-9pt}
\caption{Cross-domain generalization accuracy in image attributors. The amount of training and testing data was kept consistent across trials, and an equal number of images was sourced from MS-COCO and GPT-4 prompts for the `Train on Both' trial.}
\label{tab:cross_domain_classification}
\vspace{-12 pt}
\end{table}

\vspace{-6pt}
\subsection{Detectability of Hyperparameter Variations}
\label{sec:Analyzing_the_Detectability_of_Hyperparameter_Variations}
\vspace{-4pt}

T2I generators often have several adjustable hyperparameters at the inference stage that affect the generated image quality. A natural question that arises is whether images produced using different hyperparameters are distinguishable.
To investigate this, we target four hyperparameter choices for Stable Diffusion \cite{rombach2022highresolution_stable_diffusion}: model checkpoint, scheduler type, number of sampling steps, and initialization seed.
Specifically, we compared Stable Diffusion checkpoints 1.1 to 1.5, each of which are trained using a different number of iterations on the LAION dataset \cite{schuhmann2022laion5b}.
We then examined the detectability of images generated using eight schedulers: DDIM \cite{song2020denoising}, DDPM \cite{ho2020denoising}, Euler \cite{karras2022elucidating}, Euler with ancestral sampling \cite{karras2022elucidating}, KDPM 2 \cite{karras2022elucidating}, LMS \cite{karras2022elucidating}, PNDM \cite{karras2022elucidating}, and UniPC \cite{zhao2024unipc}.
Additionally, we generated images using both SD 2.0 and SDXL for ten different sampling steps ranging from 5 to 50, and ten different seeds ranging from 1 to 10.
For each hyperparameter choice, we train a separate EfficientFormer \cite{li2022efficientformer} to classify the generated images, and the results are illustrated in Tab. \ref{tab:detect_inference_variations} and Fig. \ref{fig:confusion_hyperparam}.
As shown in Tab. \ref{tab:detect_inference_variations}, all six classifiers can detect the hyperparameter choice better than random chance. Interestingly, the initialization seed achieves nearly 100\% accuracy, which aligns with prior work by Yu \etal \cite{yu2019attributing} that found different seeds lead to attributable GAN fingerprints. Moreover, when looking at the confusion matrix for different sampling steps using SDXL in Fig. \ref{fig:confusion_hyperparam}, we see that images generated using fewer steps are more detectable than those generated using more steps, likely because fewer steps noticeably degrades the generation quality.

\begin{table}[h!]
\vspace{-7 pt}
\centering
{\smaller %
\begin{tabularx}{\columnwidth}{c| *{3}{>{\centering\arraybackslash}X|} >{\centering\arraybackslash}X}
\toprule
           & Checkpoints & Schedulers & Steps & Seeds \\
\midrule
Random   & 20\%       & 12.5\%  & 10\%/10\%        & 10\%/10\%    \\
Accuracy     & 30.2\%       & 20.1\% & 25.9\%/56.6\%  & 98.8\%/99.9\%    \\
\bottomrule
\end{tabularx}
}
\vspace{-8pt}
\caption{Comparison of accuracy for detecting hyperparameter values based on generated images. For the `Sampling Steps' and `Seeds' trials, we trained and evaluated on images from SD 2.0 and SDXL, and the accuracies are formatted as \textit{SD 2.0 / SDXL}. Notably, the `Seeds' trial attains near perfect performance.}
\label{tab:detect_inference_variations}
\vspace{-8 pt}
\end{table}

\vspace{-6pt}
\subsection{Detectability of Post-Editing Enhancements}
\label{sec:Assessing_Detectability_Following_Post-Editing_Enhancements}
\vspace{-4pt}

\begin{figure}[!h]
    \centering
    \includegraphics[trim=0in 2.25in 0.0in 0in, clip,width=\columnwidth]{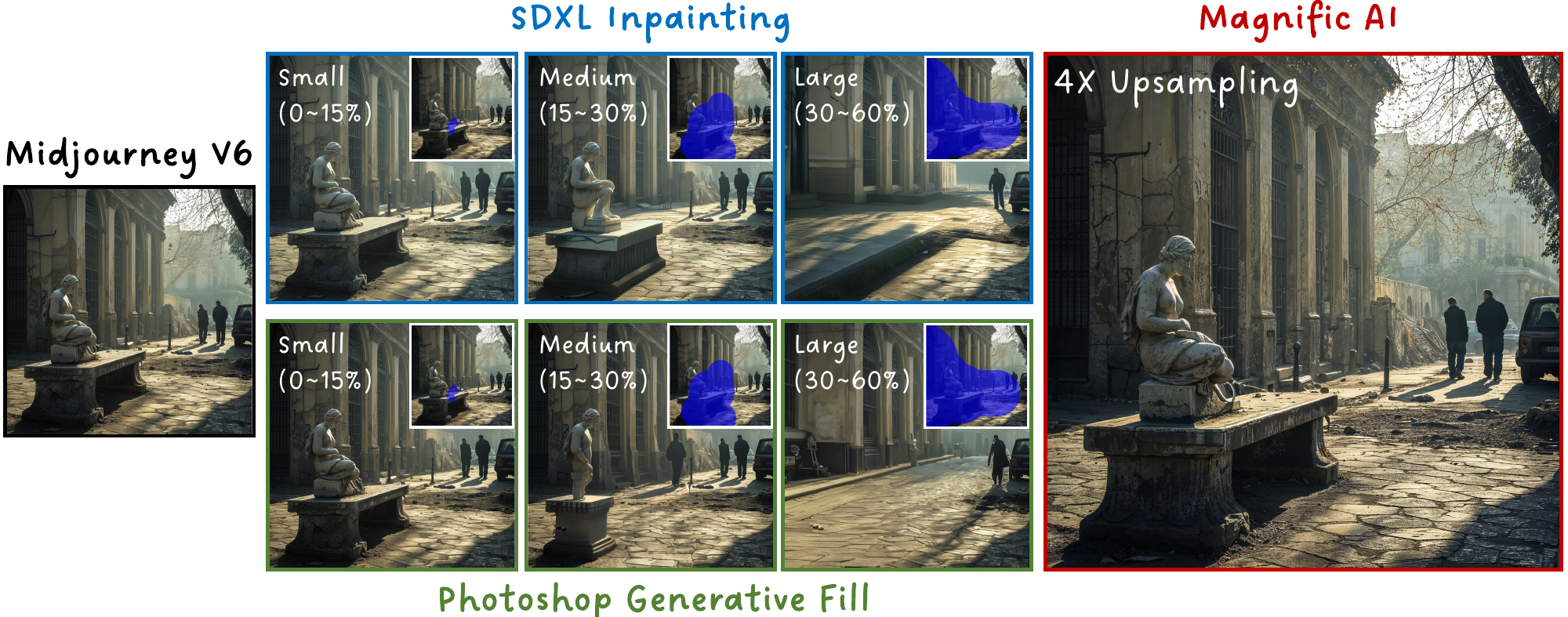}
    \vspace{-20 pt}
    \caption{\textbf{Left}: Original image generated by Midjourney 6. \textbf{Middle}: Local modifications utilizing SDXL inpainting and Photoshop Generative Fill across three masks with small, medium, and large holes. \textbf{Right}: The image upscaled 4X by Magnific AI.}
    \label{fig:post_editing}
    \vspace{-17 pt}
\end{figure}

A common workflow for utilizing AI-generated images involves users identifying unwanted artifacts or distracting areas within these images. They often import these images into additional models or software for further editing and refinement, such as using SDXL Inpainting \cite{podell2023sdxl} or Photoshop Generative Fill (Ps GenFill) \cite{Photoshop} to enhance local regions. Many T2I applications are limited to relatively low resolutions, typically around 1K, or produce images with smooth/blurry texture. Hence, some professionals upscale or refine the details of generated images using advanced tools, such as Magnific AI \cite{MagnificAI}. This practice leads to a pertinent question: Is it possible to still detect the original source generator after the images have undergone further modifications using a variety of software or other AI models? For instance, an image initially created by Midjourney 6 \cite{Midjourney} could subsequently be edited with SDXL Inpainting, Photoshop GenFill, or Magnific AI, as illustrated in Fig. \ref{fig:post_editing}.

\begin{table*}[h!]
\vspace{-24pt}
\centering
{\smaller %
\begin{tabularx}{\textwidth}{c| *{6}{>{\centering\arraybackslash}X|} >{\centering\arraybackslash}X}
\toprule
            & \multicolumn{3}{c|}{\textbf{SDXL Inpainting}} & \multicolumn{3}{c|}{\textbf{Ps Generative Fill}} & \multicolumn{1}{c}{\textbf{Magnific AI}} \\
\midrule
Edit Region Ratio   & 0 $\sim$ 15\% & 15 $\sim$ 30\% & 30 $\sim$ 60\% & 0 $\sim$ 15\% & 15 $\sim$ 30\% & 30 $\sim$ 60\% & 100\% \\
\midrule
Random Chance   & 7.69\%   & 7.69\%  & 7.69\%  & 7.69\%  & 7.69\%  & 7.69\%  & 33.33\% \\
Original Image  & 90.96\%  & 90.96\% & 90.96\% & 90.96\% & 90.96\% & 90.96\% & 93.33\% \\
Post-Editing    & 64.96\%  & 61.56\% & 55.62\% & 88.21\% & 85.44\% & 71.91\% & 70.00\% \\
\bottomrule
\end{tabularx}
}
\vspace{-8pt}
\caption{Comparison of post-editing detection accuracy across different AI models. We use the best performing image attributor in Table \ref{tab:image_attributor_metric} for evaluation, which is EfficientFormer trained with text prompts. Accuracy declines at a similar rate after modifying the image using SDXL Inpainting \cite{podell2023sdxl} and Photoshop (Ps) Generative Fill \cite{Photoshop}.}
\label{tab:post_editing_detection}
\vspace{-15 pt}
\end{table*}

To simulate typical user edits, we generated free-form masks across three size categories—small (0 to 15\%), medium (15 to 30\%), and large (30 to 60\%)—reflecting the common range of edits applied to images. These masks were applied to the entire test set for pixel regeneration using SDXL Inpainting \cite{podell2023sdxl} and Ps GenFill \cite{Photoshop}. We then assessed the best performing image attributor in Tab. \ref{tab:image_attributor_metric}, EfficientFormer trained with text prompts, on these post-edited images. According to Tab. \ref{tab:post_editing_detection}, we observed a monotonic decrease in accuracy with respect to the modified area of the images. Notably, SDXL Inpainting resulted in greater accuracy loss compared to Ps GenFill for the same images and masks. We hypothesize this disparity arises because the SDXL Inpainting model closely relates to the SDXL text-to-image (T2I) model included in our training generator pool, potentially skewing edited images towards an SDXL-like appearance, whereas Ps GenFill does not closely resemble any generator in our training set. This observation is validated in the corresponding confusion matrix, which we have shared in the supplemental materials. For texture enhancements via Magnific AI \cite{MagnificAI}, budget constraints limited our examination to 10 examples from each of the three generators: DALL-E 3, Midjourney 6, and SDXL Turbo. This limitation set a basic random chance of classification at 33.33\%. This analysis, reflected in the last column of Tab. \ref{tab:post_editing_detection}, shows approximately 23\% degradation, despite editing all pixels in the images. Despite the noted performance reductions, the accuracy for all post-edited images remains significantly above random chance, establishing a strong baseline for the task of post-editing image attribution.

\vspace{-6pt}
\section{Detecting Image Attribution Beyond RGB}
\label{sec:Detecting_Image_Attribution_Beyond_RGB}
\vspace{-4pt}

Previous studies have demonstrated that training a standard deep network can effectively distinguish between real and generated images, as well as correctly attribute generated images to their original generators. In Sec. \ref{sec:training_image_attributors}, we observed that a lightweight transformer achieves high accuracy for these tasks, mirroring these findings. These prior studies have suggested that the attributor may leverage middle-to-high frequency information to differentiate images. However, it remains unclear what exactly constitutes ``middle-to-high frequency information" and to what extent the network can still identify detectable traces in the images as we incrementally remove visual details. Thus, this section presents an extensive empirical study on the impact of progressively eliminating visual information at various levels of granularity on image attribution performance.

\begin{figure*}[!h]
 \vspace{-24 pt}
    \centering
    \includegraphics[trim=0in 0.86in 0.85in 0in, clip,width=\textwidth]{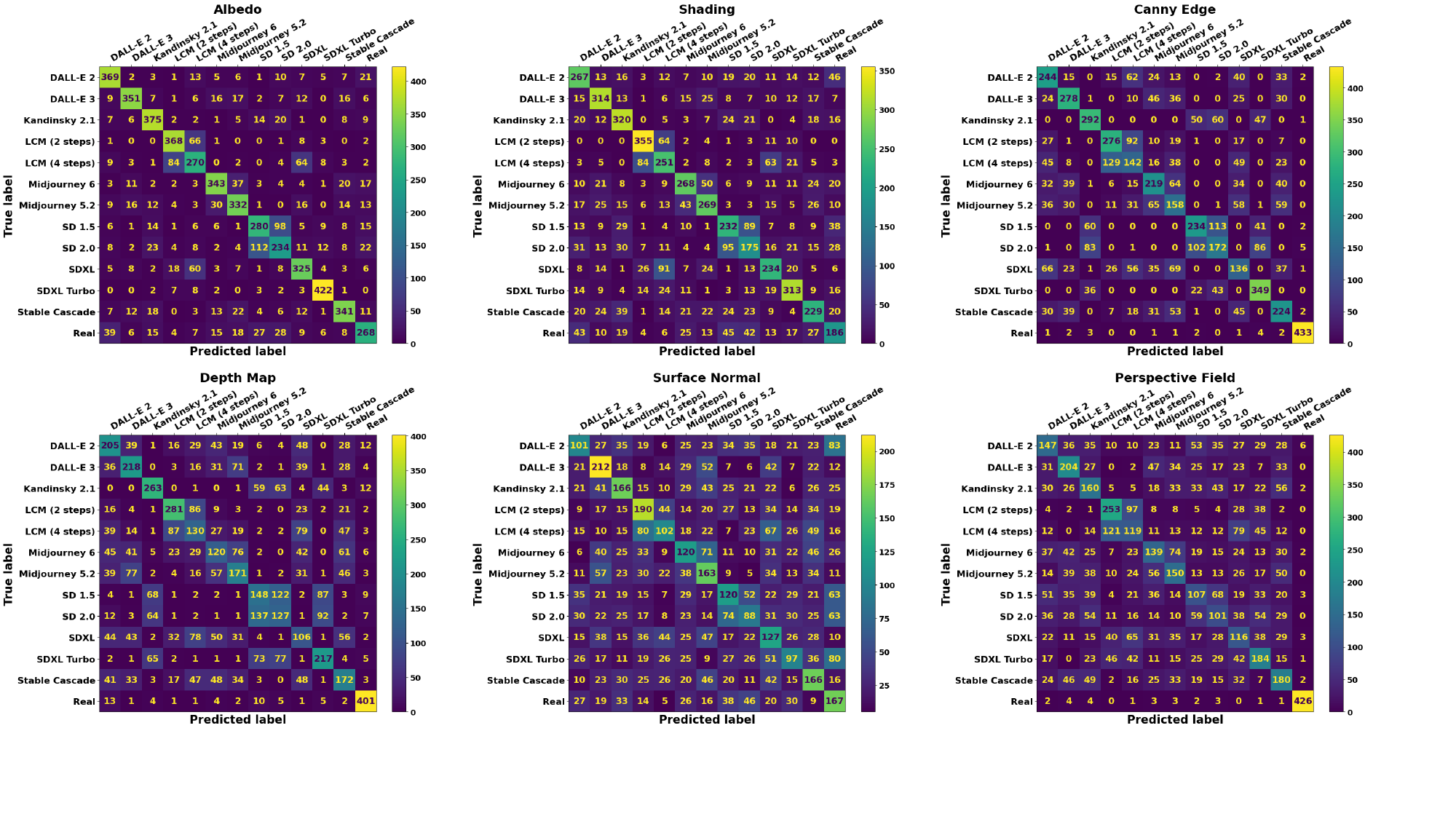}
    \vspace{-19 pt}
    \caption{Confusion matrices for image attributors trained on mid-level representations. Remarkably, attributors trained on ``Canny Edge," ``Depth Map," and ``Perspective Field" images are significantly better at detecting real images than synthetic images.}
    \label{fig:confusion_beyond_rgb}
    \vspace{-16 pt}
\end{figure*}

\begin{figure}[!h]
    \centering
    \includegraphics[trim=0in 1.35in 1.0in 0in, clip,width=\columnwidth]{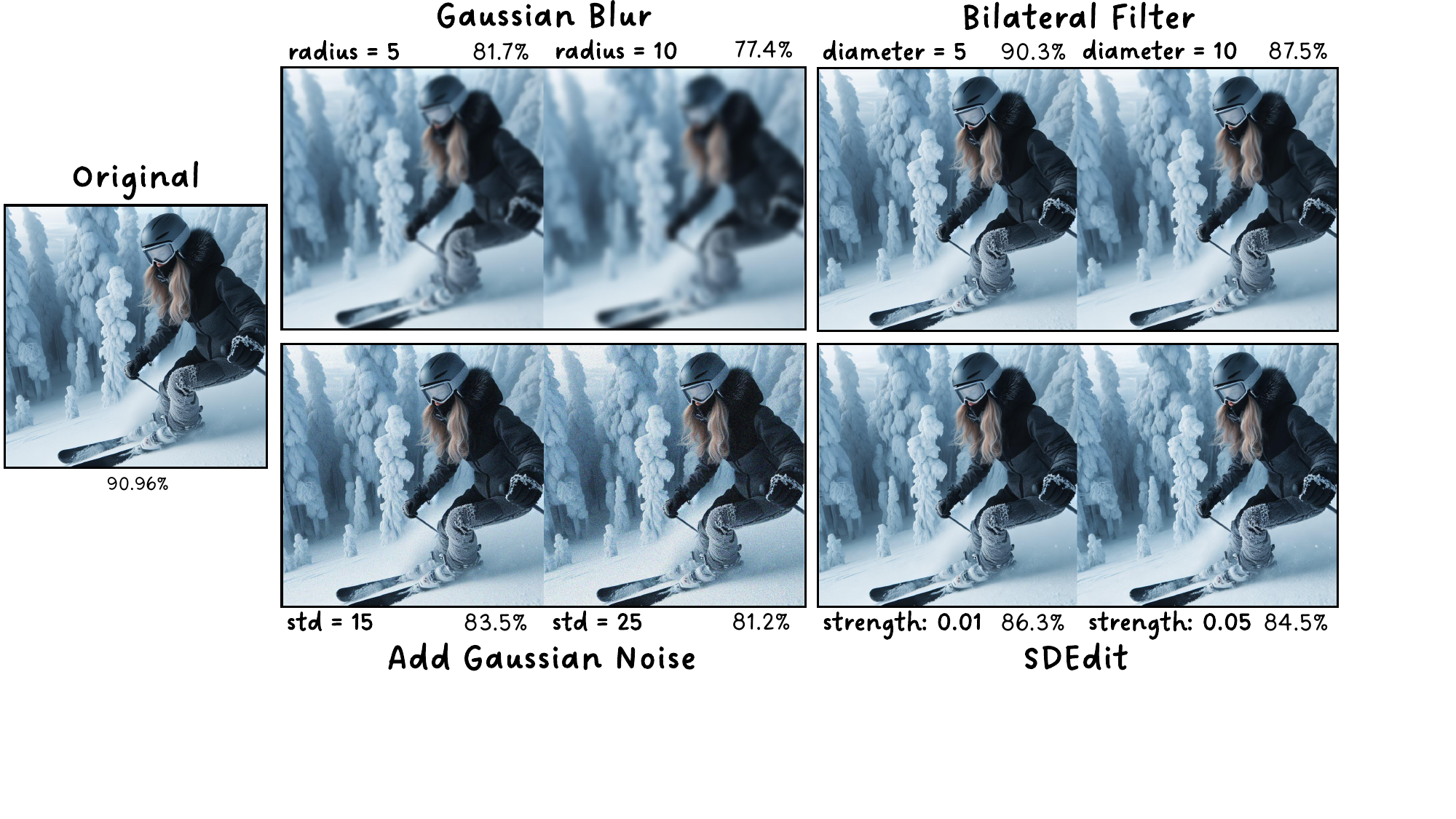}
    \vspace{-20 pt}
    \caption{We present a generated image before and after perturbing its high-frequency details. We trained EfficientFormer on images after each high-frequency perturbation and observed a mild decline in the respective test accuracy, as shown beside the images.}
    \label{fig:perturb_high_frequency}
    \vspace{-6 pt}
\end{figure}

\textbf{High-Frequency Perturbation.} Prior research \cite{corvi2023intriguing, ricker2024detection, bi2023detecting, think_twice, frequency_aware, marra2018gans, frank2020leveraging, bammey2023synthbuster, durall2020watch, dzanic2020fourier} has identified that generators leave unique fingerprints in the high-frequency domain, allowing attributors to learn these high-frequency details effectively. As an initial step, we investigate the effects of introducing high-frequency perturbations to images on the attributor's performance, which aims to enforce the classifier learn beyond high-frequency details. For simplicity, we train a separate EfficientFormer \cite{li2022efficientformer} on each set of perturbed images. Figure \ref{fig:perturb_high_frequency} illustrates our observations under four types of perturbation: Gaussian blur, bilateral filtering, adding Gaussian noise, and SDEdit \cite{meng2021sdedit}. We note that these perturbations result in a modest decrease in classification accuracy.
Specifically for SDEdit, the high-frequency traits of SDXL are embedded into every image, regardless of their source generators, by undergoing processing via the encoder, diffusion UNet, and decoder of SDXL \cite{podell2023sdxl}. Remarkably, this process led to only a minor reduction in accuracy, suggesting a robustness in the attributor's ability to identify generator-specific fingerprints despite high-frequency modifications.

\begin{figure}[!h]
 \vspace{-5 pt}
    \centering
    \includegraphics[trim=0in 4.95in 0.0in 0in, clip,width=\columnwidth]{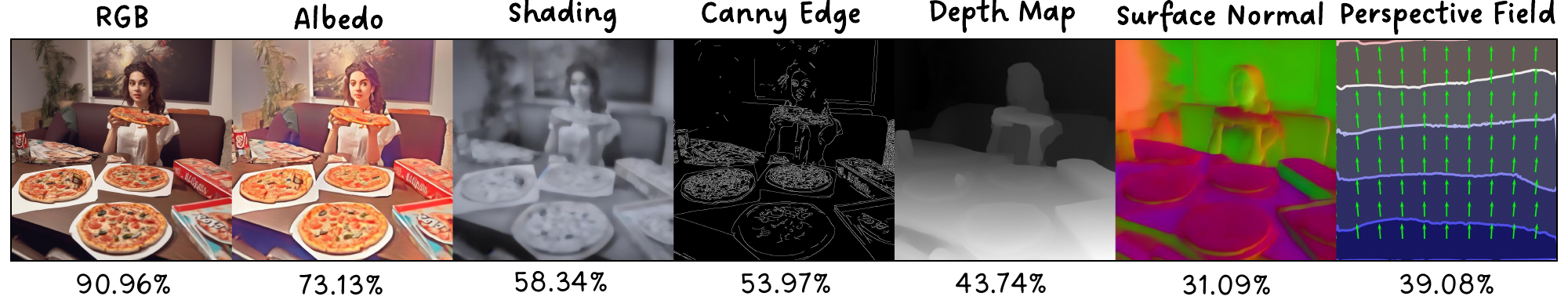}
    \vspace{-20 pt}
    \caption{We present an RGB image and its mid-level representations. We trained EfficientFormer on each mid-level representation and include the corresponding test accuracy under each image. Please keep in mind that the random chance is \textbf{$\frac{1}{13}$} or \textbf{7.69\%}. }
    \label{fig:middle_level_representation}
    \vspace{-18 pt}
\end{figure}

\textbf{Middle-Level Representation.} High-frequency perturbations result in only minor performance degradation, which suggests that the detectable traces left by different generators might also reside within the mid-frequency domain. To investigate the presence of these detectable traces, we convert the images into various mid-level representations---`Albedo' \cite{du2023generative}, `Shading' \cite{du2023generative}, `Canny Edge', `Depth Map' \cite{depthanything}, `Surface Normal' \cite{du2023generative}, and `Perspective Fields' \cite{jin2022PerspectiveFields}---utilizing readily available models. This approach aims to uncover the extent to which these mid-level frequencies carry generator-specific information that can be leveraged for attribution.
We proceed by training a distinct EfficientFormer \cite{li2022efficientformer} for each mid-level representation, and we show their classification accuracies in Fig. \ref{fig:middle_level_representation} and confusion matrices in Fig. \ref{fig:confusion_beyond_rgb}.
Notably, although the overall accuracy for the attributors trained on Canny Edge, Depth Map, and Perspective Field images is not high in Fig. \ref{fig:middle_level_representation}, they demonstrate remarkable performance at discerning real images from fake images in Fig. \ref{fig:confusion_beyond_rgb}. This finding aligns with previous work by Sarkar \etal \cite{sarkar2023shadows} suggesting that generative models often fail to generate accurate geometry.

\begin{figure*}[!th]
 \vspace{-24 pt}
    \centering
    \includegraphics[trim=0in 3.75in 0in 0in, clip,width=\textwidth]{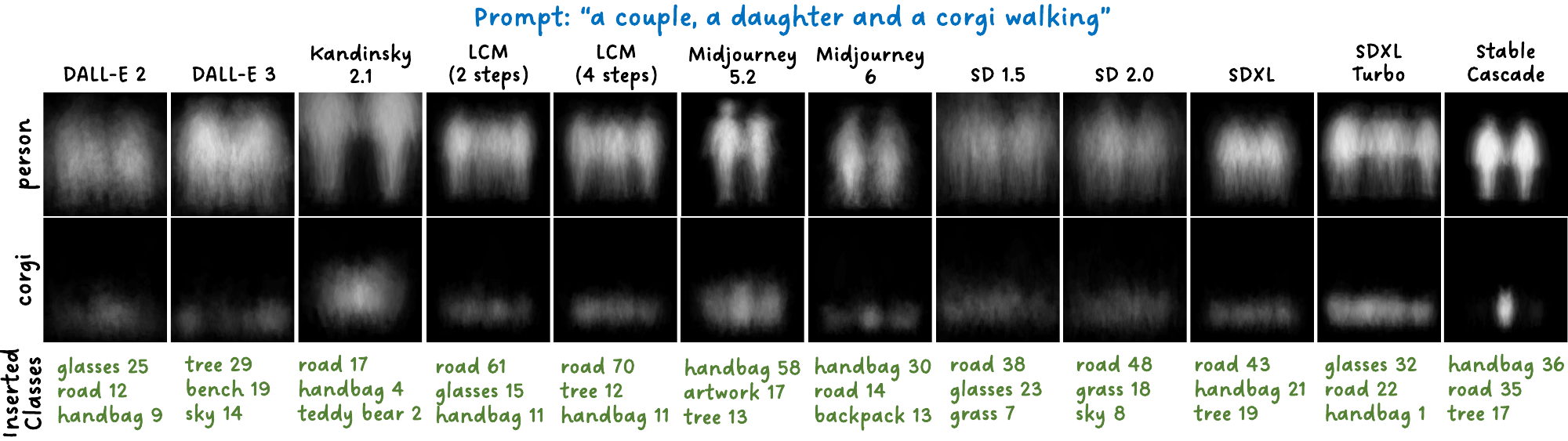}
    \vspace{-20 pt}
    \caption{Image composition analysis for a given prompt. We show the averaged segmentation masks for the `person' and `corgi' classes. Some generators put objects at specific locations. We also list the top inserted classes and how many images (out of 100) with these classes.}
    \label{fig:image_layout}
    \vspace{-15 pt}
\end{figure*}

\textbf{Image Style Representation.} Given the perceptible differences in styles or tones among image generators, it's common to observe distinct characteristics in their outputs. For instance, Midjourney \cite{Midjourney} often produces images with a `cinematic' quality, while DALL-E \cite{ramesh2022hierarchical_dalle, betker2023improving} tends to create images with overly smooth textures and cartoonish appearances, as in Fig. \ref{fig:gen_from_models_prompts}. This observation leads to a pertinent question: if we train an attributor on only stylistic representations of images, can we still identify source generators?

To capture the style representation of images, we adhere to the methodology established in prior style transfer literature \cite{johnson2016perceptual, gatys2016style_transfer}, employing a pretrained VGG network \cite{simonyan2014very} to extract features across multiple layers. Subsequently, we calculate the Gram matrix \cite{gatys2015texture} for each layer of the network. If we denote the feature at a specific layer as \(F \in \mathbb{R}^{H \times W \times N}\), then the Gram matrix is the cosine similarity between each channel in the feature representation, yielding a matrix of dimensions \(G \in \mathbb{R}^{N \times N}\). This process aims to distill the stylistic essence of images, providing a unique fingerprint for each generator's output. Specifically, we reshape and concatenate the Gram matrices extracted from multiple layers, and then train EfficientFormer \cite{li2022efficientformer} using these aggregated feature vectors.

Remarkably, the image attributor achieves an accuracy of \textbf{92.80\%} when trained on style representations, surpassing the performance of the attributor trained on original RGB images by \textbf{1.84\%}. The superior accuracy achieved by this style-based image attributor highlights the critical role of stylistic features, such as texture and color patterns, in distinguishing generators more effectively than traditional RGB data. This suggests that the unique signatures of image generators might be more intricately tied to their style rather than the direct visual content. This insight not only advances our understanding of image attribution techniques but also emphasizes the potential of leveraging stylistic elements for more nuanced AI recognition and analysis tasks.

\begin{figure}[!h]
 \vspace{-8 pt}
    \centering
    \includegraphics[trim=0in 2.3in 1.5in 0in, clip,width=\columnwidth]{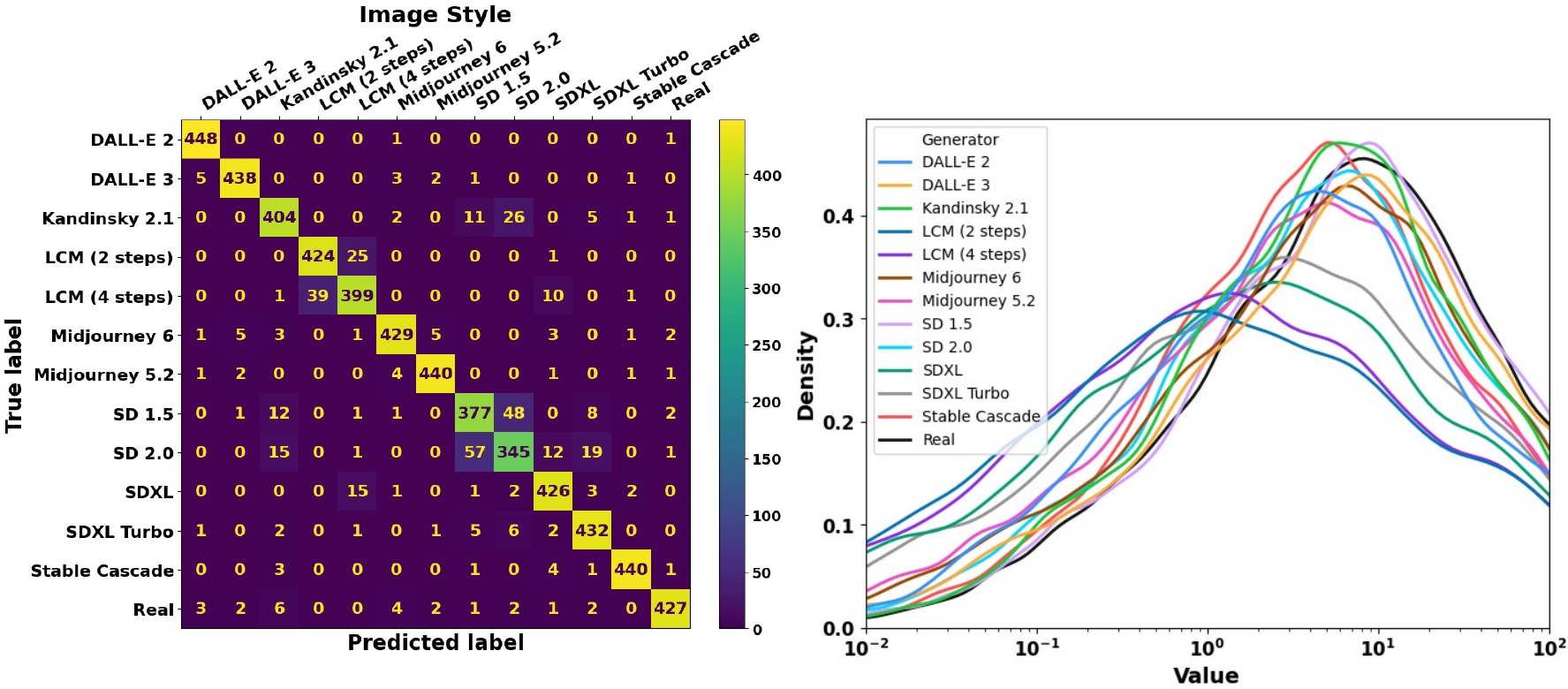}
    \vspace{-20 pt}
    \caption{\textbf{Left:} Confusion matrix for EfficientFormer trained on aggregated style features from Gram matrices. Compared to EfficientFormer trained on original RGB images in Figure \ref{fig:image_attributor_results}, training on image style reduces misclassification between generators of the same family, such as ``Midjourney 5.2 vs. Midjourney 6." \ \textbf{Right:} Density distribution of values in the averaged Gram matrix using 450 images per generator. We include real images as a distinct `generator'. Image style is moderately distinguishable across generators by analyzing Gram matrices alone. \textbf{Please zoom-in.}}
    \label{fig:image_style_representation}
    \vspace{-12 pt}
\end{figure}

Furthermore, given the exceptional performance of training on style features for image attribution, we seek to understand what insights we can extract from raw values in the Gram matrices without any model training. To achieve this, we average the Gram matrix from a single layer of VGG across 450 images per generator, and we visualize the density distribution of its values in Fig. \ref{fig:image_style_representation}. We observe that LCM (2 steps) and LCM (4 steps) have similar image style distributions, as does SDXL and SDXL Turbo. In addition, since we use real images from the MS-COCO dataset \cite{lin2015microsoft_coco_dataset}, the generators with distributions closer to that of real images in Fig. \ref{fig:image_style_representation} likely generate more natural image styles.

\textbf{Image Composition Pattern. } Beyond stylistic differences, we hypothesize that various generators might create images with unique composition patterns or layouts from the same text prompt. For instance, given identical prompts, some generators may depict humans in portrait-style photos, while others may place humans further from the camera, treating them as elements within the larger scene. These variations could stem from each generator's learning with its distinctively `curated' training data distribution and proprietary prompt augmentation techniques, features that are often integral to commercial models like DALL-E \cite{ramesh2022hierarchical_dalle, betker2023improving} and Midjourney \cite{Midjourney}. To test our hypothesis, we analyze 100 images generated from the same prompt for each generator. We employ Grounded SAM \cite{ren2024grounded} to compute segmentation masks, serving as a proxy for layout representation. For instance, as depicted in Figure \ref{fig:image_layout}, by averaging the segmentation masks for `person' and `corgi' across 100 images from each generator, created from the prompt `a couple, a daughter, and a corgi walking,' we visualize the distribution of image composition. This reveals unique layout patterns among the generators, supporting our hypothesis.

Given the noticeable variations in the layout of generated images for a specific prompt, we further investigate whether a classifier can learn to attribute images based solely on their composition. To this end, we segment 111 semantic classes using Grounded SAM \cite{ren2024grounded} and then train EfficientFormer \cite{li2022efficientformer} on the segmentation maps with their input prompts by concatenating their respective embeddings. This approach enables the classifier to achieve an accuracy of 17.66\%, despite relying on such a coarse representation. Remarkably, this accuracy is more than twice that expected by random chance (7.69\%), suggesting that distinct patterns in layout generation do indeed exist across these generators.

\vspace{-12 pt}
\section{Conclusion}
\label{sec:conclusion}
\vspace{-10 pt}

In this study, we present in-depth analyses on the detection and attribution of images generated by modern text-to-image (T2I) diffusion models. Through rigorous testing, our image attributors, trained to recognize outputs from 12 different T2I diffusion models along with a category for real images, reached an impressive accuracy of over 90\%, significantly surpassing random chance. Our investigation into the role of text prompts, the challenge of distinguishing generators within the same family, and the ability to generalize across domains provides comprehensive insights. Pioneeringly, we delved into the detectability of hyperparameter adjustments at inference time and assessed the effects of post-editing on attribution accuracy. Going beyond mere RGB analysis, we introduce a new framework for identifying detectable traces across various levels of visual detail, offering profound insights into the underlying mechanics of image attribution. These analyses provide fresh perspectives on image forensics aimed at alleviating the threat of synthetic images on copyright protection and digital forgery.

\textbf{Acknowledgements.}
This work is supported by funds provided by the National Science Foundation and by DoD OUSD (R\&E) under Cooperative Agreement PHY-2229929 (The NSF AI Institute for Artificial and Natural Intelligence).

{\small
\bibliographystyle{ieee_fullname}
\bibliography{egbib}

\begin{thebibliography}{10}\itemsep=-1pt

\bibitem{achiam2023gpt}
Josh Achiam, Steven Adler, Sandhini Agarwal, Lama Ahmad, Ilge Akkaya, Florencia~Leoni Aleman, Diogo Almeida, Janko Altenschmidt, Sam Altman, Shyamal Anadkat, et~al.
\newblock Gpt-4 technical report.
\newblock {\em arXiv preprint arXiv:2303.08774}, 2023.

\bibitem{AdobeFirefly}
Adobe firefly.
\newblock \url{https://firefly.adobe.com/inspire/images}.

\bibitem{amoroso2023parents}
Roberto Amoroso, Davide Morelli, Marcella Cornia, Lorenzo Baraldi, Alberto~Del Bimbo, and Rita Cucchiara.
\newblock Parents and children: Distinguishing multimodal deepfakes from natural images, 2023.

\bibitem{bammey2023synthbuster}
Quentin Bammey.
\newblock Synthbuster: Towards detection of diffusion model generated images.
\newblock {\em IEEE Open Journal of Signal Processing}, PP:1--9, 01 2023.

\bibitem{betker2023improving}
James Betker, Gabriel Goh, Li Jing, Tim Brooks, Jianfeng Wang, Linjie Li, Long Ouyang, Juntang Zhuang, Joyce Lee, Yufei Guo, et~al.
\newblock Improving image generation with better captions.
\newblock {\em Computer Science. https://cdn. openai. com/papers/dall-e-3. pdf}, 2(3):8, 2023.

\bibitem{bi2023detecting}
Xiuli Bi, Bo Liu, Fan Yang, Bin Xiao, Weisheng Li, Gao Huang, and Pamela~C. Cosman.
\newblock Detecting generated images by real images only, 2023.

\bibitem{bird2023cifake}
Jordan~J. Bird and Ahmad Lotfi.
\newblock Cifake: Image classification and explainable identification of ai-generated synthetic images, 2023.

\bibitem{bui2022repmix}
Tu Bui, Ning Yu, and John Collomosse.
\newblock Repmix: Representation mixing for robust attribution of synthesized images, 2022.

\bibitem{chai2020makes}
Lucy Chai, David Bau, Ser-Nam Lim, and Phillip Isola.
\newblock What makes fake images detectable? understanding properties that generalize, 2020.

\bibitem{chen2018learning}
Chen Chen, Qifeng Chen, Jia Xu, and Vladlen Koltun.
\newblock Learning to see in the dark, 2018.

\bibitem{chen2024single}
Jiaxuan Chen, Jieteng Yao, and Li Niu.
\newblock A single simple patch is all you need for ai-generated image detection, 2024.

\bibitem{chen2023surface}
Yida Chen, Fernanda Viégas, and Martin Wattenberg.
\newblock Beyond surface statistics: Scene representations in a latent diffusion model, 2023.

\bibitem{2023mmpretrain}
MMPreTrain Contributors.
\newblock Openmmlab's pre-training toolbox and benchmark.
\newblock \url{https://github.com/open-mmlab/mmpretrain}, 2023.

\bibitem{corvi2023intriguing}
Riccardo Corvi, Davide Cozzolino, Giovanni Poggi, Koki Nagano, and Luisa Verdoliva.
\newblock Intriguing properties of synthetic images: from generative adversarial networks to diffusion models, 2023.

\bibitem{corvi2022detection}
Riccardo Corvi, Davide Cozzolino, Giada Zingarini, Giovanni Poggi, Koki Nagano, and Luisa Verdoliva.
\newblock On the detection of synthetic images generated by diffusion models, 2022.

\bibitem{cozzolino2023raising}
Davide Cozzolino, Giovanni Poggi, Riccardo Corvi, Matthias Nießner, and Luisa Verdoliva.
\newblock Raising the bar of ai-generated image detection with clip, 2023.

\bibitem{dai2019second}
Tao Dai, Jianrui Cai, Yongbing Zhang, Shu-Tao Xia, and Lei Zhang.
\newblock Second-order attention network for single image super-resolution.
\newblock In {\em Proceedings of the IEEE Conference on Computer Vision and Pattern Recognition}, pages 11065--11074, 2019.

\bibitem{think_twice}
Chengdong Dong, Ajay Kumar, and Eryun Liu.
\newblock Think twice before detecting gan-generated fake images from their spectral domain imprints.
\newblock In {\em 2022 IEEE/CVF Conference on Computer Vision and Pattern Recognition (CVPR)}, pages 7855--7864, 2022.

\bibitem{du2023generative}
Xiaodan Du, Nicholas Kolkin, Greg Shakhnarovich, and Anand Bhattad.
\newblock Generative models: What do they know? do they know things? let's find out!, 2023.

\bibitem{durall2020watch}
Ricard Durall, Margret Keuper, and Janis Keuper.
\newblock Watch your up-convolution: Cnn based generative deep neural networks are failing to reproduce spectral distributions, 2020.

\bibitem{dzanic2020fourier}
Tarik Dzanic, Karan Shah, and Freddie Witherden.
\newblock Fourier spectrum discrepancies in deep network generated images, 2020.

\bibitem{epstein2023online}
David~C. Epstein, Ishan Jain, Oliver Wang, and Richard Zhang.
\newblock Online detection of ai-generated images.
\newblock In {\em ICCV DeepFake Analysis and Detection Workshop}, 2023.

\bibitem{frank2020leveraging}
Joel Frank, Thorsten Eisenhofer, Lea Schönherr, Asja Fischer, Dorothea Kolossa, and Thorsten Holz.
\newblock Leveraging frequency analysis for deep fake image recognition, 2020.

\bibitem{gatys2015texture}
Leon~A. Gatys, Alexander~S. Ecker, and Matthias Bethge.
\newblock Texture synthesis using convolutional neural networks, 2015.

\bibitem{gatys2016style_transfer}
Leon~A. Gatys, Alexander~S. Ecker, and Matthias Bethge.
\newblock Image style transfer using convolutional neural networks.
\newblock In {\em 2016 IEEE Conference on Computer Vision and Pattern Recognition (CVPR)}, pages 2414--2423, 2016.

\bibitem{jacobgilpytorchcam}
Jacob Gildenblat and contributors.
\newblock Pytorch library for cam methods.
\newblock \url{https://github.com/jacobgil/pytorch-grad-cam}, 2021.

\bibitem{girish2021discovery}
Sharath Girish, Saksham Suri, Saketh Rambhatla, and Abhinav Shrivastava.
\newblock Towards discovery and attribution of open-world gan generated images, 2021.

\bibitem{goodfellow2014generative}
Ian~J. Goodfellow, Jean Pouget-Abadie, Mehdi Mirza, Bing Xu, David Warde-Farley, Sherjil Ozair, Aaron Courville, and Yoshua Bengio.
\newblock Generative adversarial networks, 2014.

\bibitem{gragnaniello2021gan}
Diego Gragnaniello, Davide Cozzolino, Francesco Marra, Giovanni Poggi, and Luisa Verdoliva.
\newblock Are gan generated images easy to detect? a critical analysis of the state-of-the-art, 2021.

\bibitem{guarnera2023level}
Luca Guarnera, Oliver Giudice, and Sebastiano Battiato.
\newblock Level up the deepfake detection: a method to effectively discriminate images generated by gan architectures and diffusion models, 2023.

\bibitem{guo2023hierarchical}
Xiao Guo, Xiaohong Liu, Zhiyuan Ren, Steven Grosz, Iacopo Masi, and Xiaoming Liu.
\newblock Hierarchical fine-grained image forgery detection and localization, 2023.

\bibitem{ha2024organic}
Anna Yoo~Jeong Ha, Josephine Passananti, Ronik Bhaskar, Shawn Shan, Reid Southen, Haitao Zheng, and Ben~Y. Zhao.
\newblock Organic or diffused: Can we distinguish human art from ai-generated images?, 2024.

\bibitem{ho2020denoising}
Jonathan Ho, Ajay Jain, and Pieter Abbeel.
\newblock Denoising diffusion probabilistic models.
\newblock {\em Advances in neural information processing systems}, 33:6840--6851, 2020.

\bibitem{hu2021lora}
Edward~J. Hu, Yelong Shen, Phillip Wallis, Zeyuan Allen-Zhu, Yuanzhi Li, Shean Wang, Lu Wang, and Weizhu Chen.
\newblock Lora: Low-rank adaptation of large language models, 2021.

\bibitem{jin2022PerspectiveFields}
Linyi Jin, Jianming Zhang, Yannick Hold-Geoffroy, Oliver Wang, Kevin Matzen, Matthew Sticha, and David~F. Fouhey.
\newblock Perspective fields for single image camera calibration.
\newblock {\em CVPR}, 2023.

\bibitem{johnson2016perceptual}
Justin Johnson, Alexandre Alahi, and Li Fei-Fei.
\newblock Perceptual losses for real-time style transfer and super-resolution.
\newblock In {\em Computer Vision--ECCV 2016: 14th European Conference, Amsterdam, The Netherlands, October 11-14, 2016, Proceedings, Part II 14}, pages 694--711. Springer, 2016.

\bibitem{ju2022fusing}
Yan Ju, Shan Jia, Lipeng Ke, Hongfei Xue, Koki Nagano, and Siwei Lyu.
\newblock Fusing global and local features for generalized ai-synthesized image detection, 2022.

\bibitem{karras2022elucidating}
Tero Karras, Miika Aittala, Timo Aila, and Samuli Laine.
\newblock Elucidating the design space of diffusion-based generative models.
\newblock {\em Advances in Neural Information Processing Systems}, 35:26565--26577, 2022.

\bibitem{laszkiewicz2023single}
Mike Laszkiewicz, Jonas Ricker, Johannes Lederer, and Asja Fischer.
\newblock Single-model attribution of generative models through final-layer inversion.
\newblock {\em arXiv preprint arXiv:2306.06210}, 2023.

\bibitem{li2022efficientformer}
Yanyu Li, Geng Yuan, Yang Wen, Ju Hu, Georgios Evangelidis, Sergey Tulyakov, Yanzhi Wang, and Jian Ren.
\newblock Efficientformer: Vision transformers at mobilenet speed.
\newblock {\em Advances in Neural Information Processing Systems}, 35:12934--12949, 2022.

\bibitem{lin2015microsoft_coco_dataset}
Tsung-Yi Lin, Michael Maire, Serge Belongie, Lubomir Bourdev, Ross Girshick, James Hays, Pietro Perona, Deva Ramanan, C.~Lawrence Zitnick, and Piotr Dollár.
\newblock Microsoft coco: Common objects in context, 2015.

\bibitem{lin2014microsoft}
Tsung-Yi Lin, Michael Maire, Serge Belongie, James Hays, Pietro Perona, Deva Ramanan, Piotr Doll{\'a}r, and C~Lawrence Zitnick.
\newblock Microsoft coco: Common objects in context.
\newblock In {\em Computer Vision--ECCV 2014: 13th European Conference, Zurich, Switzerland, September 6-12, 2014, Proceedings, Part V 13}, pages 740--755. Springer, 2014.

\bibitem{liu2024model}
Fengyuan Liu, Haochen Luo, Yiming Li, Philip Torr, and Jindong Gu.
\newblock Model-agnostic origin attribution of generated images with few-shot examples.
\newblock {\em arXiv preprint arXiv:2404.02697}, 2024.

\bibitem{liu2020global}
Zhengzhe Liu, Xiaojuan Qi, and Philip Torr.
\newblock Global texture enhancement for fake face detection in the wild, 2020.

\bibitem{loshchilov2019decoupled_adamw}
Ilya Loshchilov and Frank Hutter.
\newblock Decoupled weight decay regularization, 2019.

\bibitem{luo2023latent_consistency_model_lcm}
Simian Luo, Yiqin Tan, Longbo Huang, Jian Li, and Hang Zhao.
\newblock Latent consistency models: Synthesizing high-resolution images with few-step inference, 2023.

\bibitem{MagnificAI}
Magnific ai.
\newblock \url{https://magnific.ai}.

\bibitem{marra2018gans}
Francesco Marra, Diego Gragnaniello, Luisa Verdoliva, and Giovanni Poggi.
\newblock Do gans leave artificial fingerprints?, 2018.

\bibitem{marra2019incremental}
Francesco Marra, Cristiano Saltori, Giulia Boato, and Luisa Verdoliva.
\newblock Incremental learning for the detection and classification of gan-generated images, 2019.

\bibitem{exploit_visual}
Falko Matern, Christian Riess, and Marc Stamminger.
\newblock Exploiting visual artifacts to expose deepfakes and face manipulations.
\newblock In {\em 2019 IEEE Winter Applications of Computer Vision Workshops (WACVW)}, pages 83--92, 2019.

\bibitem{meng2021sdedit}
Chenlin Meng, Yutong He, Yang Song, Jiaming Song, Jiajun Wu, Jun-Yan Zhu, and Stefano Ermon.
\newblock Sdedit: Guided image synthesis and editing with stochastic differential equations.
\newblock {\em arXiv preprint arXiv:2108.01073}, 2021.

\bibitem{MetaAI}
Meta ai imagine.
\newblock \url{https://www.meta.ai/}.

\bibitem{Midjourney}
Midjourney.
\newblock \url{https://www.midjourney.com}.

\bibitem{ojha2023universal}
Utkarsh Ojha, Yuheng Li, and Yong~Jae Lee.
\newblock Towards universal fake image detectors that generalize across generative models, 2023.

\bibitem{oquab2023dinov2}
Maxime Oquab, Timoth{\'e}e Darcet, Th{\'e}o Moutakanni, Huy Vo, Marc Szafraniec, Vasil Khalidov, Pierre Fernandez, Daniel Haziza, Francisco Massa, Alaaeldin El-Nouby, et~al.
\newblock Dinov2: Learning robust visual features without supervision.
\newblock {\em arXiv preprint arXiv:2304.07193}, 2023.

\bibitem{pernias2023wuerstchen}
Pablo Pernias, Dominic Rampas, Mats~L. Richter, Christopher~J. Pal, and Marc Aubreville.
\newblock Wuerstchen: An efficient architecture for large-scale text-to-image diffusion models, 2023.

\bibitem{Photoshop}
Generative fill - ai image filler - adobe photoshop.
\newblock \url{https://www.adobe.com/products/photoshop/generative-fill.html}.

\bibitem{podell2023sdxl}
Dustin Podell, Zion English, Kyle Lacey, Andreas Blattmann, Tim Dockhorn, Jonas M{\"u}ller, Joe Penna, and Robin Rombach.
\newblock Sdxl: Improving latent diffusion models for high-resolution image synthesis.
\newblock {\em arXiv preprint arXiv:2307.01952}, 2023.

\bibitem{radford2021learning}
Alec Radford, Jong~Wook Kim, Chris Hallacy, Aditya Ramesh, Gabriel Goh, Sandhini Agarwal, Girish Sastry, Amanda Askell, Pamela Mishkin, Jack Clark, et~al.
\newblock Learning transferable visual models from natural language supervision.
\newblock In {\em International conference on machine learning}, pages 8748--8763. PMLR, 2021.

\bibitem{ramesh2022hierarchical_dalle}
Aditya Ramesh, Prafulla Dhariwal, Alex Nichol, Casey Chu, and Mark Chen.
\newblock Hierarchical text-conditional image generation with clip latents, 2022.

\bibitem{razzhigaev2023kandinsky}
Anton Razzhigaev, Arseniy Shakhmatov, Anastasia Maltseva, Vladimir Arkhipkin, Igor Pavlov, Ilya Ryabov, Angelina Kuts, Alexander Panchenko, Andrey Kuznetsov, and Denis Dimitrov.
\newblock Kandinsky: an improved text-to-image synthesis with image prior and latent diffusion, 2023.

\bibitem{ren2024grounded}
Tianhe Ren, Shilong Liu, Ailing Zeng, Jing Lin, Kunchang Li, He Cao, Jiayu Chen, Xinyu Huang, Yukang Chen, Feng Yan, et~al.
\newblock Grounded sam: Assembling open-world models for diverse visual tasks.
\newblock {\em arXiv preprint arXiv:2401.14159}, 2024.

\bibitem{ricker2024detection}
Jonas Ricker, Simon Damm, Thorsten Holz, and Asja Fischer.
\newblock Towards the detection of diffusion model deepfakes, 2024.

\bibitem{rombach2022highresolution_stable_diffusion}
Robin Rombach, Andreas Blattmann, Dominik Lorenz, Patrick Esser, and Björn Ommer.
\newblock High-resolution image synthesis with latent diffusion models, 2022.

\bibitem{sarkar2023shadows}
Ayush Sarkar, Hanlin Mai, Amitabh Mahapatra, Svetlana Lazebnik, D.~A. Forsyth, and Anand Bhattad.
\newblock Shadows don't lie and lines can't bend! generative models don't know projective geometry...for now, 2023.

\bibitem{sauer2023adversarial}
Axel Sauer, Dominik Lorenz, Andreas Blattmann, and Robin Rombach.
\newblock Adversarial diffusion distillation.
\newblock {\em arXiv preprint arXiv:2311.17042}, 2023.

\bibitem{schuhmann2022laion5b}
Christoph Schuhmann, Romain Beaumont, Richard Vencu, Cade Gordon, Ross Wightman, Mehdi Cherti, Theo Coombes, Aarush Katta, Clayton Mullis, Mitchell Wortsman, Patrick Schramowski, Srivatsa Kundurthy, Katherine Crowson, Ludwig Schmidt, Robert Kaczmarczyk, and Jenia Jitsev.
\newblock Laion-5b: An open large-scale dataset for training next generation image-text models, 2022.

\bibitem{Selvaraju_2019_gradcam}
Ramprasaath~R. Selvaraju, Michael Cogswell, Abhishek Das, Ramakrishna Vedantam, Devi Parikh, and Dhruv Batra.
\newblock Grad-cam: Visual explanations from deep networks via gradient-based localization.
\newblock {\em International Journal of Computer Vision}, 128(2):336–359, Oct. 2019.

\bibitem{sha2023defake}
Zeyang Sha, Zheng Li, Ning Yu, and Yang Zhang.
\newblock De-fake: Detection and attribution of fake images generated by text-to-image generation models, 2023.

\bibitem{simonyan2014very}
Karen Simonyan and Andrew Zisserman.
\newblock Very deep convolutional networks for large-scale image recognition.
\newblock {\em arXiv preprint arXiv:1409.1556}, 2014.

\bibitem{sinitsa2023deep}
Sergey Sinitsa and Ohad Fried.
\newblock Deep image fingerprint: Towards low budget synthetic image detection and model lineage analysis, 2023.

\bibitem{song2020denoising}
Jiaming Song, Chenlin Meng, and Stefano Ermon.
\newblock Denoising diffusion implicit models.
\newblock {\em arXiv preprint arXiv:2010.02502}, 2020.

\bibitem{StableDiffusion3}
Stable diffusion 3.
\newblock \url{https://stability.ai/news/stable-diffusion-3}.

\bibitem{tan2023learning}
Chuangchuang Tan, Yao Zhao, Shikui Wei, Guanghua Gu, and Yunchao Wei.
\newblock Learning on gradients: Generalized artifacts representation for gan-generated images detection.
\newblock In {\em Proceedings of the IEEE/CVF Conference on Computer Vision and Pattern Recognition}, pages 12105--12114, 2023.

\bibitem{frequency_aware}
Cheng Tian, Zhiming Luo, Guimin Shi, and Shaozi Li.
\newblock Frequency-aware attentional feature fusion for deepfake detection.
\newblock In {\em ICASSP 2023 - 2023 IEEE International Conference on Acoustics, Speech and Signal Processing (ICASSP)}, pages 1--5, 2023.

\bibitem{von-platen-etal-2022-diffusers}
Patrick von Platen, Suraj Patil, Anton Lozhkov, Pedro Cuenca, Nathan Lambert, Kashif Rasul, Mishig Davaadorj, and Thomas Wolf.
\newblock Diffusers: State-of-the-art diffusion models.
\newblock \url{https://github.com/huggingface/diffusers}, 2022.

\bibitem{wang2020cnngenerated}
Sheng-Yu Wang, Oliver Wang, Richard Zhang, Andrew Owens, and Alexei~A. Efros.
\newblock Cnn-generated images are surprisingly easy to spot... for now, 2020.

\bibitem{wang2023dire}
Zhendong Wang, Jianmin Bao, Wengang Zhou, Weilun Wang, Hezhen Hu, Hong Chen, and Houqiang Li.
\newblock Dire for diffusion-generated image detection, 2023.

\bibitem{wang2024did}
Zhenting Wang, Chen Chen, Yi Zeng, Lingjuan Lyu, and Shiqing Ma.
\newblock Where did i come from? origin attribution of ai-generated images.
\newblock {\em Advances in neural information processing systems}, 36, 2024.

\bibitem{wu2023generalizable}
Haiwei Wu, Jiantao Zhou, and Shile Zhang.
\newblock Generalizable synthetic image detection via language-guided contrastive learning, 2023.

\bibitem{depthanything}
Lihe Yang, Bingyi Kang, Zilong Huang, Xiaogang Xu, Jiashi Feng, and Hengshuang Zhao.
\newblock Depth anything: Unleashing the power of large-scale unlabeled data.
\newblock In {\em CVPR}, 2024.

\bibitem{yu2019attributing}
Ning Yu, Larry Davis, and Mario Fritz.
\newblock Attributing fake images to gans: Learning and analyzing gan fingerprints, 2019.

\bibitem{yu2021artificial}
Ning Yu, Vladislav Skripniuk, Sahar Abdelnabi, and Mario Fritz.
\newblock Artificial fingerprinting for generative models: Rooting deepfake attribution in training data.
\newblock In {\em Proceedings of the IEEE/CVF International conference on computer vision}, pages 14448--14457, 2021.

\bibitem{zhan2023does_sd}
Guanqi Zhan, Chuanxia Zheng, Weidi Xie, and Andrew Zisserman.
\newblock What does stable diffusion know about the 3d scene?, 2023.

\bibitem{zhang2023diffusion}
Yichi Zhang and Xiaogang Xu.
\newblock Diffusion noise feature: Accurate and fast generated image detection, 2023.

\bibitem{zhao2024unipc}
Wenliang Zhao, Lujia Bai, Yongming Rao, Jie Zhou, and Jiwen Lu.
\newblock Unipc: A unified predictor-corrector framework for fast sampling of diffusion models.
\newblock {\em Advances in Neural Information Processing Systems}, 36, 2024.

\bibitem{zhong2024patchcraft}
Nan Zhong, Yiran Xu, Sheng Li, Zhenxing Qian, and Xinpeng Zhang.
\newblock Patchcraft: Exploring texture patch for efficient ai-generated image detection, 2024.

\bibitem{zhu2023gendet}
Mingjian Zhu, Hanting Chen, Mouxiao Huang, Wei Li, Hailin Hu, Jie Hu, and Yunhe Wang.
\newblock Gendet: Towards good generalizations for ai-generated image detection, 2023.

\end{thebibliography}
}

\newpage
\appendix

\vspace{-20pt}
\section{Human Performance}
\vspace{-5pt}

In computer vision and machine learning, human performance is typically seen as the benchmark for AI models. However, in the case of image attribution, the scenario reverses—AI significantly outperforms humans. This is highlighted by an experiment conducted by one of our co-authors, who has extensive experience with AI-generated images. Tasked with attributing 650 images to their correct source generators, the co-author achieved only a \textbf{37.23\%} accuracy rate. This figure, while better than the 7.69\% random chance level, is markedly inferior to the accuracy of our top AI classifier that has 90\%+ accuracy. This outcome underlines the exceptional challenge of image attribution, where even well-informed individuals struggle. It shows the necessity of AI in assisting humans with tasks that are beyond their natural proficiency, emphasizing AI's potential to enhance human performance in specialized domains.

From the perspective of the human evaluator, differentiating between certain AI image generators and others can be nuanced yet discernible. The Latent Consistency Models (LCM) \cite{luo2023latent_consistency_model_lcm}, at 2 and 4 steps, are notable for their occasional oversmooth artifacts, a result of undersampling, making them easier to identify compared to other models. DALL-E 3 \cite{betker2023improving} is distinguished by its tendency to produce surreal, cartoonish images, though these often exhibit repetitive patterns. DALL-E 2 \cite{ramesh2022hierarchical_dalle}, on the other hand, is characterized by a unique `sharp' visual artifact, likely a consequence of its pixel diffusion process in the decoder, setting it apart from other models. Midjourney versions 5.2 and 6 \cite{Midjourney} typically deliver the highest quality images, sometimes with a distinctive cinematic style.

Real images, however, are more straightforward to identify. One can often look at the detailed object regions—like hands and text—where AI-generated images tend to falter. The naturalistic photo style of real images also serves as a key differentiation factor from AI-generated content. Other generators, such as SD 1.5 \cite{rombach2022highresolution_stable_diffusion}, SD 2.0 \cite{rombach2022highresolution_stable_diffusion}, SDXL \cite{podell2023sdxl}, SDXL Turbo \cite{sauer2023adversarial}, Kandinsky 2.1 \cite{razzhigaev2023kandinsky}, and Stable Cascade \cite{pernias2023wuerstchen}, present a greater challenge for human evaluators to distinguish due to the subtlety of their differences.

\vspace{-5pt}
\section{Data and Implementation Details}
\vspace{-5pt}

\textbf{GPT-4 Generated Prompts.} 
Building upon Section \nonarxiv{3}\arxiv{\ref{sec:dataset}} of our main paper, this section describes the methodology behind generating creative and surreal prompts using GPT-4 \cite{achiam2023gpt}. As illustrated in Fig. \ref{fig:gpt4_prompt} in the supplemental, our process begins with the formulation of system-level instructions directing GPT-4 to act as an assistant for writing text prompts. We then supply a specific context and a collection of several hundred exemplary prompts. This setup enables GPT-4 to synthesize and generate new, innovative prompts based on the provided examples and context.

\begin{figure*}[!th]
 \vspace{-20 pt}
    \centering
    \includegraphics[trim=0in 1.0in 0.0in 0in, clip,width=\textwidth]{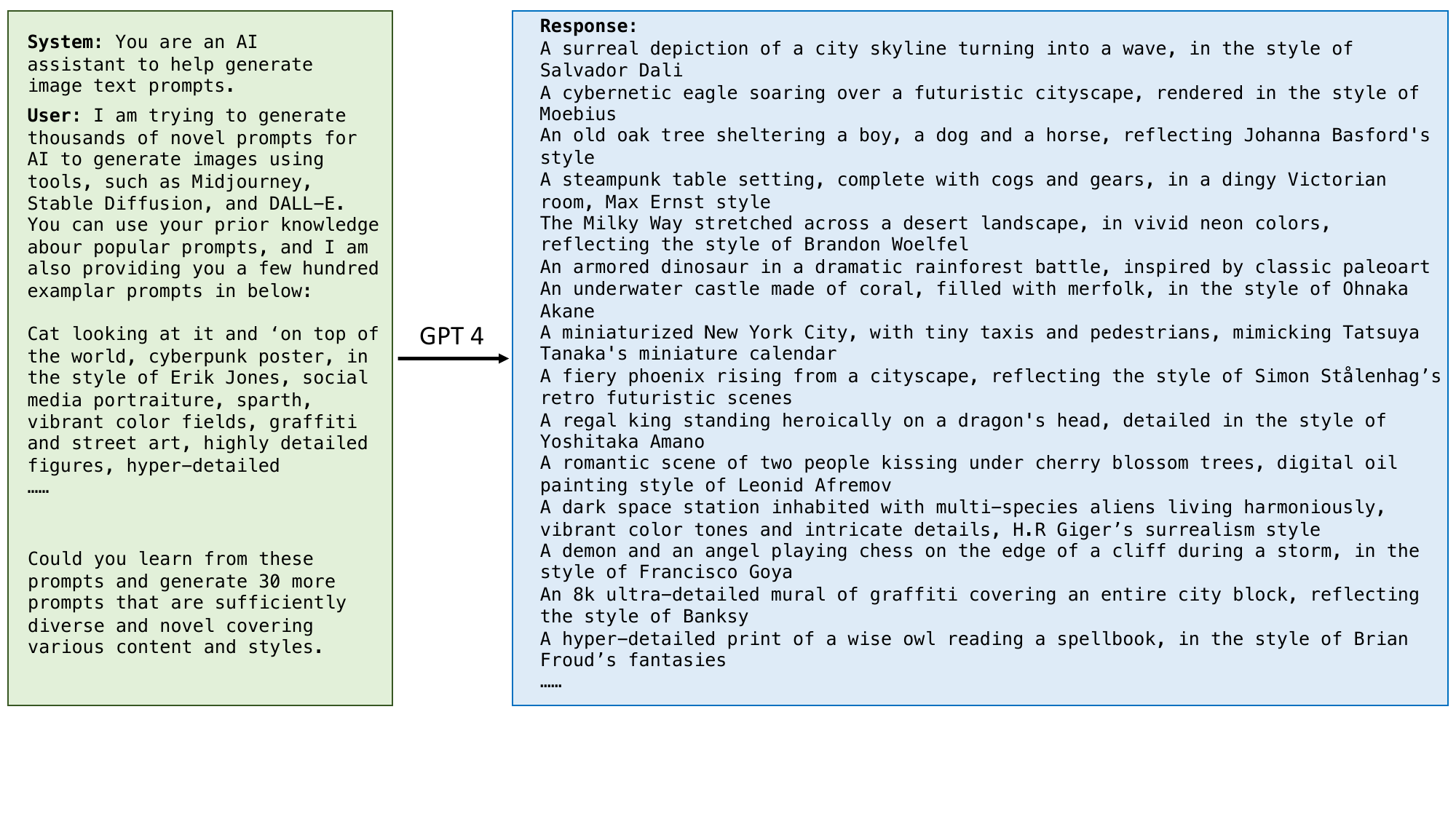}
    \vspace{-19 pt}
    \caption{An illustration of how we use the GPT-4 API to massively generate thousands of creative and surreal prompts. }
    \label{fig:gpt4_prompt}
    \vspace{-2 pt}
\end{figure*}

\textbf{Image Generation.}
We employed 12 T2I diffusion models to generate RGB images without watermarks, and the generated image sizes are as follows:
\begin{itemize}
    \renewcommand{\labelitemi}{\LARGE$\cdot$}
    \item \textbf{512 $\times$ 512:} Kandinsky 2.1, SD 1.1, SD 1.2, SD 1.3, SD 1.4, SD 1.5, SD 2.0, SDXL Turbo
    \item \textbf{1024 $\times$ 1024:} DALL-E 2, DALL-E 3, LCM (2 steps), LCM (4 steps), Midjourney 5.2, Midjourney 6, SDXL, Stable Cascade
\end{itemize}
\vspace{-2pt}
We also use 5000 real images from the MS-COCO \cite{lin2015microsoft_coco_dataset} 2017 validation set.

\textbf{More Visualizations of Hyperparameter Variations.}
As an extension of Fig. 2 in the main paper, we show more image generations with hyperparameter variations in Fig. \ref{fig:gen_from_hyperparameter_variations_supp} in the supplemental. 

\begin{figure*}[!h]
 \vspace{-5 pt}
    \centering
    \includegraphics[trim=0in 1.3in 0in 0in, clip,width=\textwidth]{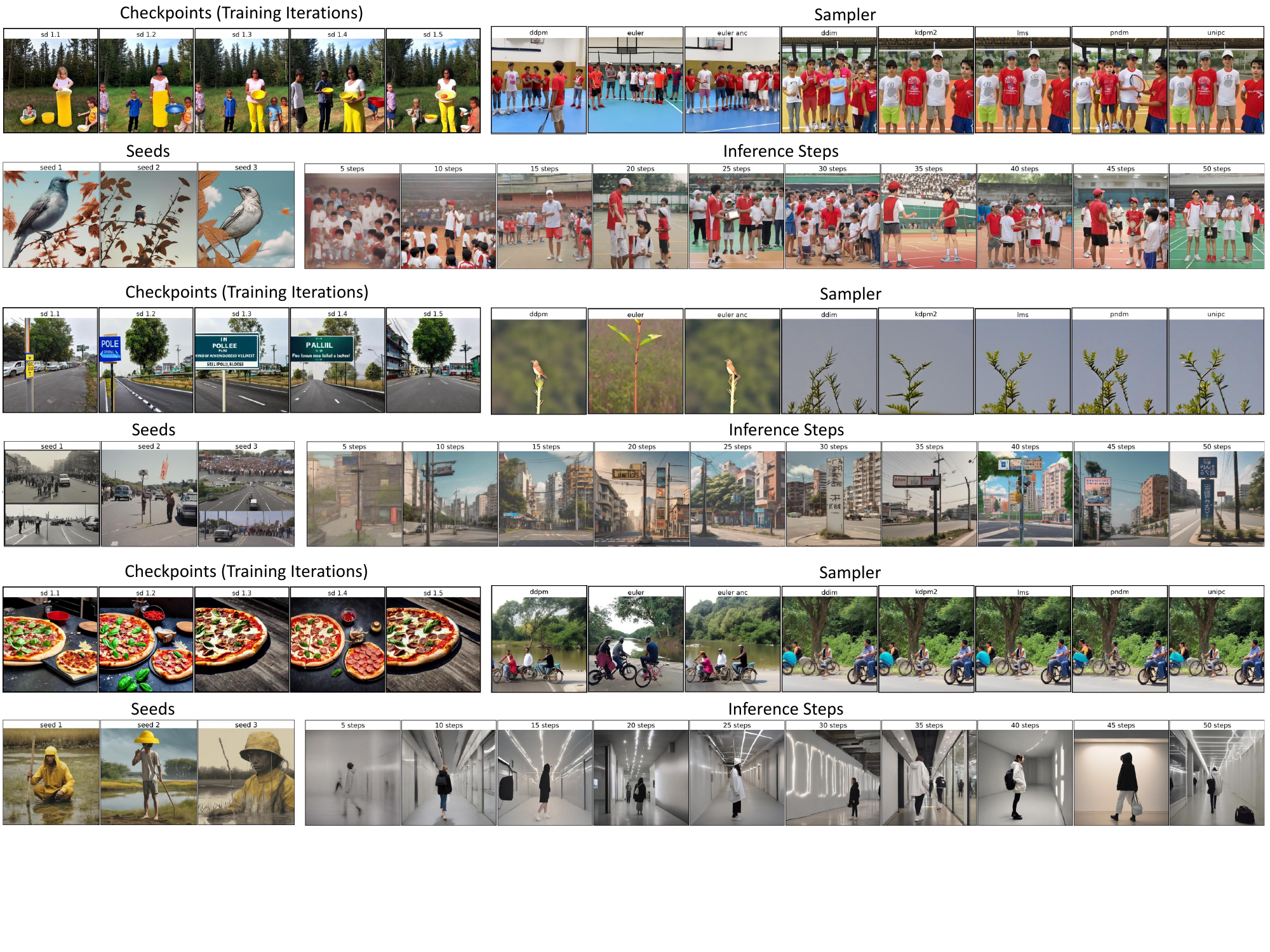}
    \vspace{-20 pt}
    \caption{More examples showcasing the diversity in generated images influenced by varying hyperparameters: different model checkpoints within the same architecture, diverse scheduling algorithms, varied initialization seeds, and a range of inference steps.}
    \label{fig:gen_from_hyperparameter_variations_supp}
    \vspace{-15 pt}
\end{figure*}

\textbf{Training Data.}
For Sections \nonarxiv{4.1}\arxiv{\ref{sec:training_image_attributors}} and \nonarxiv{5}\arxiv{\ref{sec:Detecting_Image_Attribution_Beyond_RGB}} in the main paper, we view image attribution as a 13-way classification task with 12 text-to-image diffusion models and 1 set of real images. An exception is the cross-domain generalization study, where we exclude real images as a 13$\textsuperscript{th}$ class because there are no real images for the GPT-4 generated prompts. It's important to note that we use 3200 training, 450 validation, and 450 testing images per class.

For Section \nonarxiv{4.2}\arxiv{\ref{sec:Analyzing_the_Detectability_of_Hyperparameter_Variations}}, we analyze four hyperparameters: Stable Diffusion checkpoint, scheduler type, number of sampling steps, and initialization seed. When training classifiers for SD checkpoints, schedulers, and sampling steps, we use 20000 training, 2500 validation, and 2500 testing images per class. For seeds, we use 3200 training, 450 validation, and 450 testing images per class.

For Section \nonarxiv{4.3}\arxiv{\ref{sec:Assessing_Detectability_Following_Post-Editing_Enhancements}}, we run inference using the EfficientFormer \cite{li2022efficientformer} trained with text prompts from Sec. \nonarxiv{4.1}\arxiv{\ref{sec:training_image_attributors}}. For SDXL Inpainting \cite{podell2023sdxl} and Photoshop Generative Fill \cite{Photoshop}, we use 450 images from each of the 13 classes. For Magnific AI \cite{MagnificAI}, we use 10 images from each of DALL-E 3, Midjourney 6, and SDXL Turbo.

\textbf{Data Augmentation.} During training, we first resize each image to have a shorter edge of size $224$ using bicubic interpolation, then center crop the image to size $224 \times 224$, and finally randomly flip the image horizontally with probability $0.5$. During validation and testing, we only resize and center crop the images.

\textbf{Image Attributors.} We selected three network architectures for the image attribution task, and we use the code implementation from MMPretrain \cite{2023mmpretrain}. Our primary architecture is EfficientFormer-L3 \cite{li2022efficientformer} trained from scratch because it is a lightweight transformer.
Moreover, we employ a pretrained, frozen transformer backbone attached to a linear probe (LP) or multilayer perceptron (MLP). The backbone is either CLIP ViT-B/16 \cite{radford2021learning} or DINOv2 ViT-L/14 \cite{oquab2023dinov2}, and the MLP consists of three linear layers with sigmoid activation and hidden dimension 256. For the linear probe and MLP classifier heads, there are 768 channels in the input feature map for CLIP+LP and CLIP+MLP, and 1024 channels for DINOv2+LP and DINOv2+MLP.

To train image attributors with text prompts, we compute text embeddings using a pretrained CLIP \cite{radford2021learning} text encoder. Then, we concatenate image embeddings from the backbone with text embeddings as input to the classifier head.

For all image attributors, we set a batch size of 128 and train for 2000 epochs. We use the checkpoint with the best validation accuracy. Additionally, we utilize the AdamW optimizer \cite{loshchilov2019decoupled_adamw} with learning rate 0.0002 and weight decay 0.05. The learning rate scheduler has a linear warm-up period of 20 epochs, followed by a cosine annealing schedule with a minimum learning rate of 0.00001.

\textbf{Perspective Fields.} We use the code implementation from \cite{jin2022PerspectiveFields}. Each input to the attributor trained on Perspective Fields has a size of $640 \times 640 \times 3$. The first $640 \times 640$ channel contains latitude values, and the next two $640 \times 640$ channels contain gravity values. We adapt the code from \cite{jin2022PerspectiveFields} to visualize the Perspective Field on a black image in Fig. \nonarxiv{7}\arxiv{\ref{fig:post_editing}} of the main paper.

\textbf{How Gram Matrix Relates to Image Style.}
Gatys \etal \cite{gatys2015texture} characterize the texture of an image by computing correlations between feature channels in each layer of a convolutional neural network. These correlations are given by the Gram matrix, which is the inner product of vectorized feature maps. Extending their method to image style, Gatys \etal \cite{gatys2016style_transfer} incorporate feature correlations, \ie Gram matrices, from multiple layers of the network to obtain a multi-scale representation of the image that extracts texture details without the global arrangement. Intuitively, employing different layers of the network leads to style representations at varying scales because features capture more complex information in later network layers. Thus, we aggregate Gram matrices from three layers of a pretrained VGG network to train our image attributor on image style representations.

\textbf{Adapting to New Text-to-Image Diffusion Models.}
Our work provides a seamless integration pathway for new generative models. For instance, to incorporate a new generator such as SD 2.0, one would simply generate approximately 5,000 images, add them to the existing training dataset, and retrain the models. This process typically requires around three days using a single RTX 4090 GPU. We intend to continually update our image attributor to include popular new open-source generators. Moreover, should there be a model not yet incorporated, anyone could replicate this integration process independently, as we plan to release all related code and datasets to the community.

\vspace{-5pt}
\section{Additional Experiments}
\label{sec:add_experiments}
\vspace{-5pt}

\subsection{Color Analysis}
\vspace{-3pt}

\begin{figure*}[!h]
 \vspace{-20 pt}
    \centering
    \includegraphics[trim=0in 0in 6.35in 0in, clip,width=\textwidth]{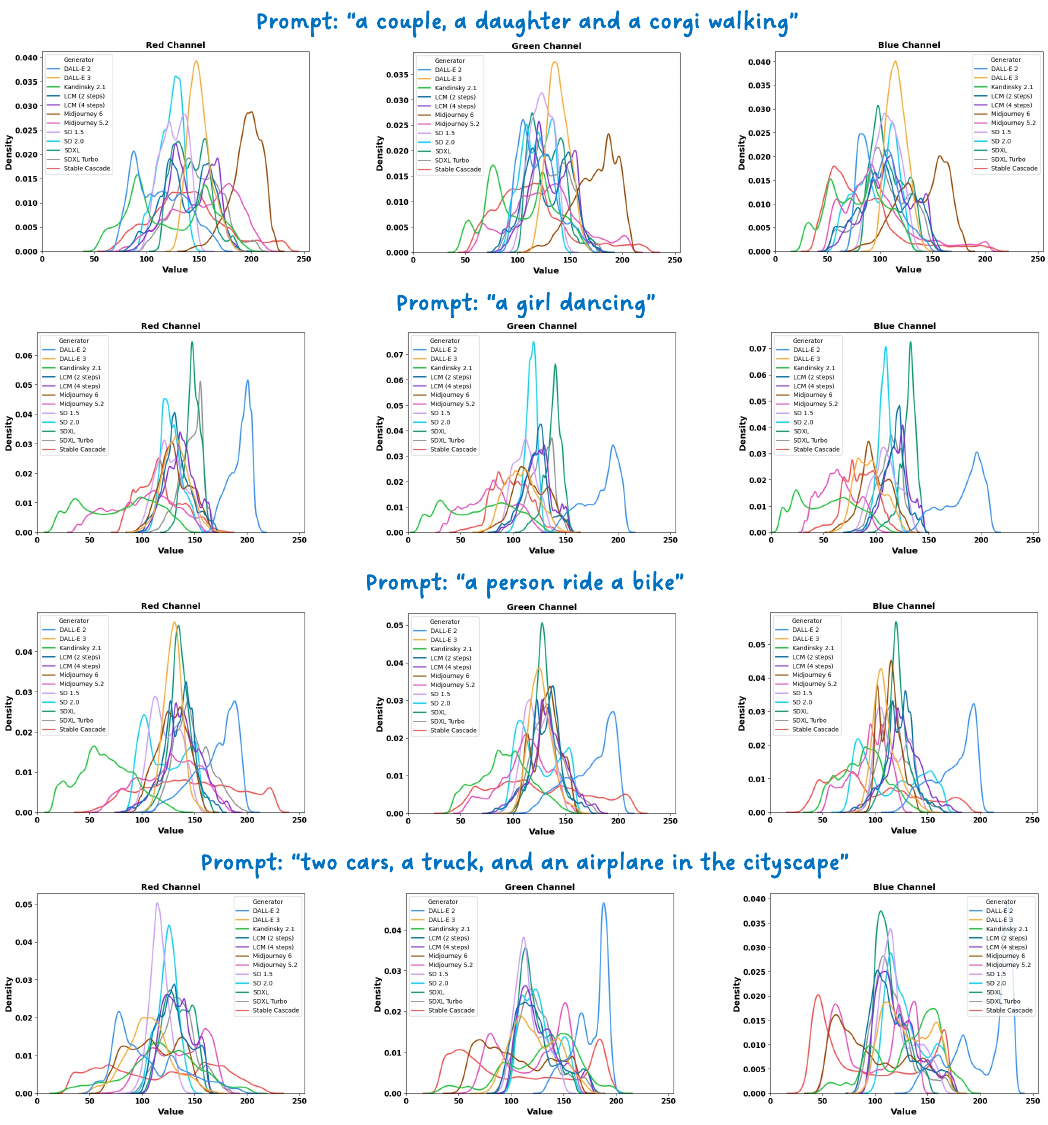}
    \vspace{-20 pt}
    \caption{Density distribution of pixel values in RGB color channels after averaging 100 images for each prompt and generator. Kandinsky 2.1 \cite{razzhigaev2023kandinsky}, Midjourney 5.2 \cite{Midjourney}, and Stable Cascade \cite{pernias2023wuerstchen} tend to create images covering a wider range of pixel intensities.}
    \label{fig:color_distribution}
    \vspace{-15 pt}
\end{figure*}

\begin{figure*}[!h]
 \vspace{-20 pt}
    \centering
    \includegraphics[trim=0in 1.5in 0.1in 0in, clip,width=\textwidth]{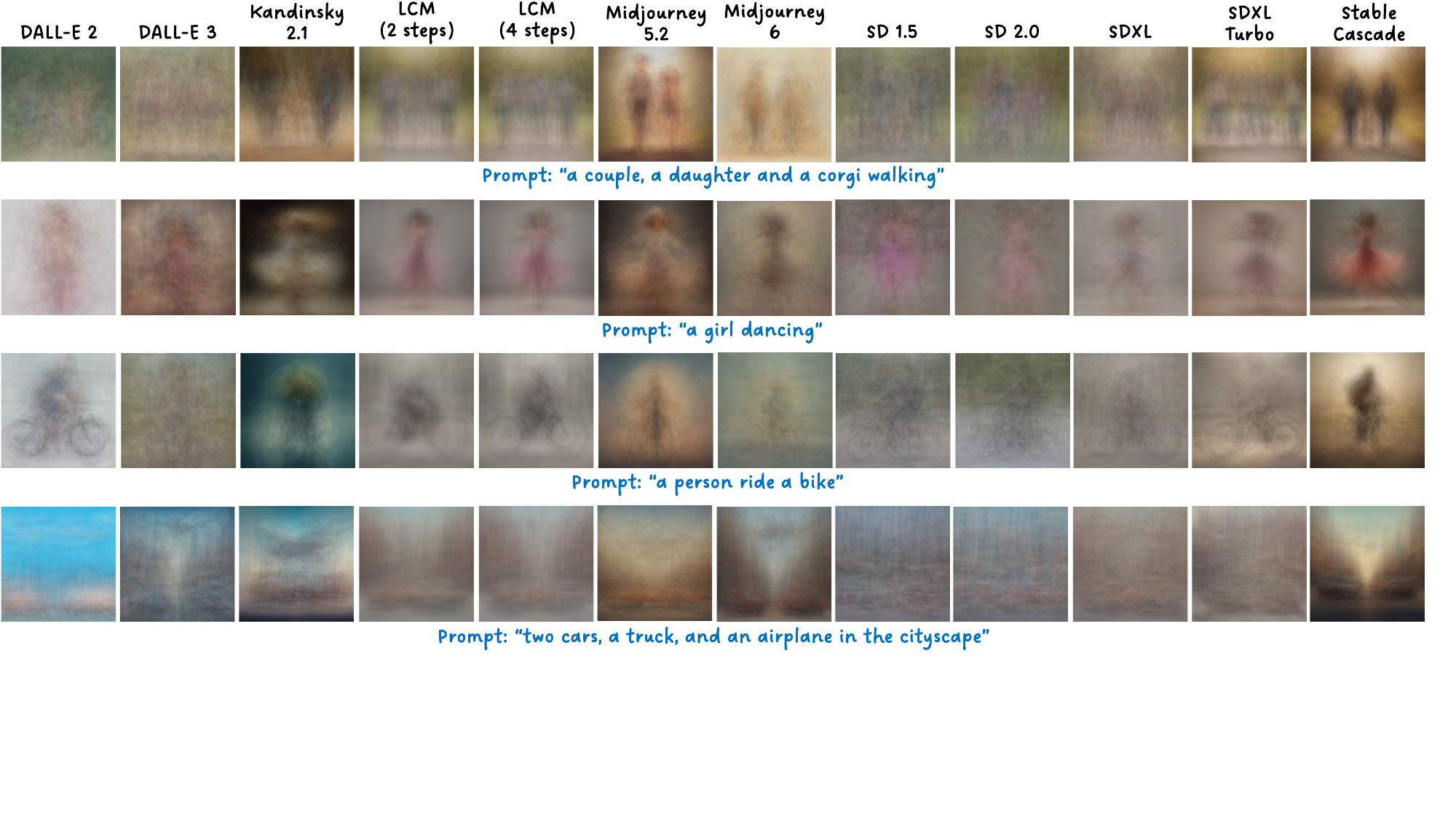}
    \vspace{-20 pt}
    \caption{Visualization of 100 images averaged together for each prompt and generator. Consistent with our observations in Fig. \ref{fig:color_distribution}, we see that Kandinsky 2.1 \cite{razzhigaev2023kandinsky}, Midjourney 5.2 \cite{Midjourney}, and Stable Cascade \cite{pernias2023wuerstchen} often produce images with glow and shadow effects.}
    \label{fig:color_images}
    \vspace{-5 pt}
\end{figure*}

In addition to studying image style and image composition pattern, we examine whether different generators produce images with distinct color schemes. We use 100 images generated from a set of fixed prompts for our analysis. In Fig. \ref{fig:color_distribution}, we visualize the density distribution of pixel values in each RGB color channel. We discover that Kandinsky 2.1 \cite{razzhigaev2023kandinsky}, Midjourney 5.2 \cite{Midjourney}, and Stable Cascade \cite{pernias2023wuerstchen} often generate images with a wider range of pixel intensity values. In Fig. \ref{fig:color_images}, we observe that these three generators often create images with glow and shadow effects, which can lead to higher and lower intensities.

\vspace{-5pt}
\subsection{Comparison of Frozen vs. Fine-tuned CLIP/ DINOv2 Backbone}
\vspace{-3pt}

In Section \nonarxiv{4.1}\arxiv{\ref{sec:training_image_attributors}} of the main paper, we evaluated the accuracy of a frozen CLIP \cite{radford2021learning} backbone connected with a linear probe and MLP, and a frozen DINOv2 \cite{oquab2023dinov2} backbone with a similar configuration. In this section, we compare using a frozen and fine-tuned backbone for the CLIP and DINOv2 linear probes. Table \ref{tab:frozen_finetuned_clip_dinov2} indicates that a CLIP backbone provides marginally better performance than a DINOv2 backbone when the backbone is frozen. However, the reverse holds true when the backbone is fine-tuned.

\begin{table*}[h!]
\centering
\begin{tabularx}{\textwidth}{l| *{3}{>{\centering\arraybackslash}X|} >{\centering\arraybackslash}X}
\toprule
    & \multicolumn{2}{c|}{CLIP + LP}    & \multicolumn{2}{c}{DINOv2 + LP} \\
\midrule
Backbone  &  Frozen   & Fine-tuned  &  Frozen  & Fine-tuned  \\
\midrule
Accuracy  &  70.15\%  & 95.31\%   &  67.68\%  &  96.67\%      \\
Precision &  69.95\%  & 95.51\%   &  67.36\%  &  96.71\%      \\
Recall    &  70.15\%  & 95.32\%   &  67.68\%  &  96.67\%      \\
F1        &  70.00\%  & 95.34\%   &  67.45\%  &  96.67\%      \\
\bottomrule
\end{tabularx}
\vspace{-7pt}
\caption{Quantitative comparison of using a frozen or fine-tuned backbone to train CLIP \cite{radford2021learning} and DINOv2 \cite{oquab2023dinov2} linear probes. CLIP achieves higher accuracy than DINOv2 when the backbone is frozen, but the opposite is true when the backbone is fine-tuned.}
\label{tab:frozen_finetuned_clip_dinov2}
\vspace{-5 pt}
\end{table*}

\vspace{-5pt}
\arxiv{\vspace{-20pt}}
\subsection{Image Resolutions}
\vspace{-3pt}

\begin{figure*}[!th]
    \centering
    \includegraphics[trim=0in 2.15in 0.1in 0in, clip,width=\textwidth]{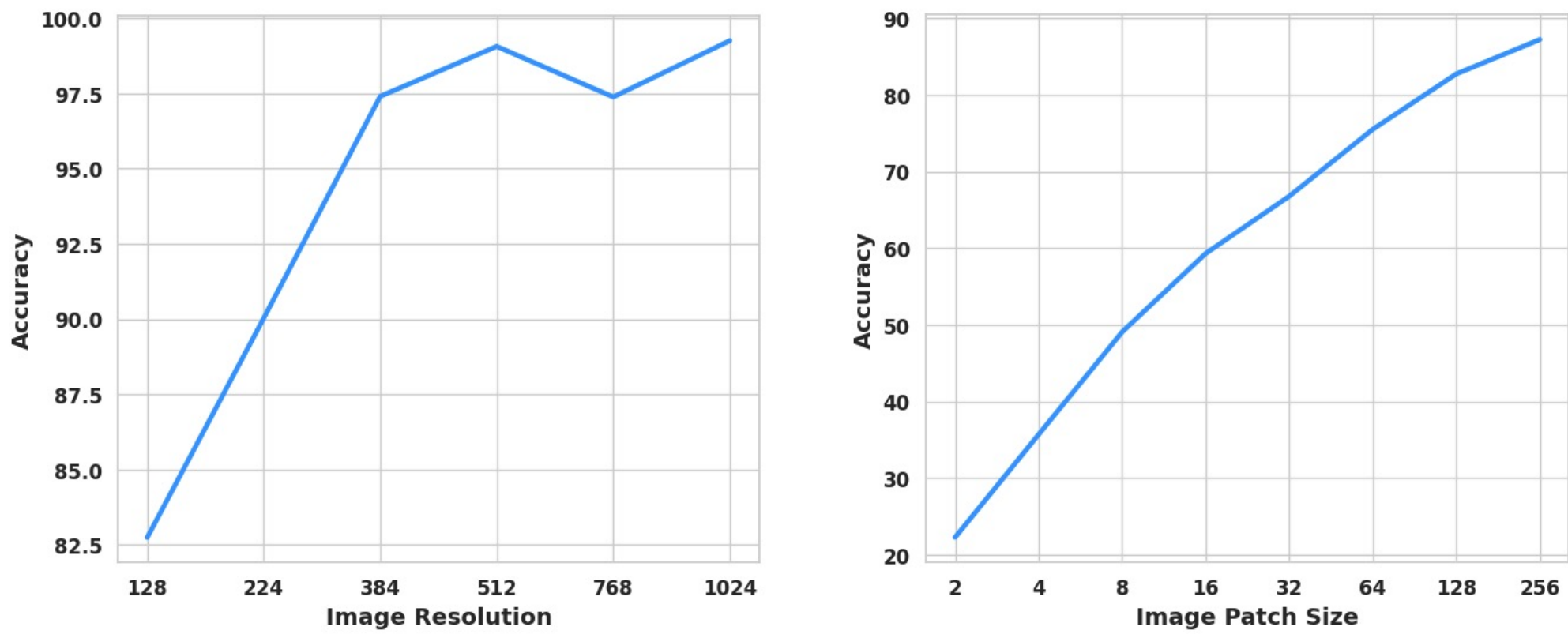}
    \vspace{-17pt}
    \caption{\textbf{Left:} Accuracy of our EfficientFormer \cite{li2022efficientformer} image attributor across six image resolutions on the 13-way classification task. In general, accuracy increases as image resolution increases. \textbf{Right:} Accuracy of EfficientFormer across eight image patch sizes. Interestingly, using $2 \times 2$ image patches can achieve 22.29\% accuracy, whereas the probability of randomly guessing the correct generator is $\frac{1}{13}$, corresponding to 7.69\%.}
    \label{fig:image_res_patch}
    \vspace{-15 pt}
\end{figure*}

The default EfficientFormer \cite{li2022efficientformer} takes inputs of size $224 \times 224$. We examine the performance of using five additional image resolutions between $128 \times 128$ and $1024 \times 1024$ for image attribution. As illustrated on the left side of Fig. \ref{fig:image_res_patch}, accuracy tends to increase as image resolution increases.

\vspace{-5pt}
\subsection{Cropped Image Patches}
\vspace{-5pt}

Our previous experiments use most, if not all, image pixels for the image attribution task. We also explore the opposite: how few pixels are necessary to achieve good performance? Inspired by \cite{zhong2024patchcraft, chen2024single}, we crop a single patch of each image and then train EfficientFormer \cite{li2022efficientformer} on these patches instead of the full-sized images. Specifically, we first resize each original image to have a shorter edge of size $512$, then center crop the image to create a patch of size $k \times k$, and finally resize the patch to $224 \times 224$. We utilized $k = [2, 4, 8, 16, 32, 64, 128, 256]$ and resized images using bicubic interpolation.
On the right side of Fig. \ref{fig:image_res_patch}, we see that accuracy increases with image patch size. Remarkably, even training an image attributor on $2 \times 2$ patches can lead to 22.29\% accuracy, which is well above the random chance accuracy of 7.69\%.

\vspace{-5pt}
\subsection{Potential Application of Model Stealing}
\begin{figure}[!ht]
    \vspace{-5 pt}
    \centering
    \includegraphics[trim=0in 0in 2.6in 0in, clip,width=\columnwidth]{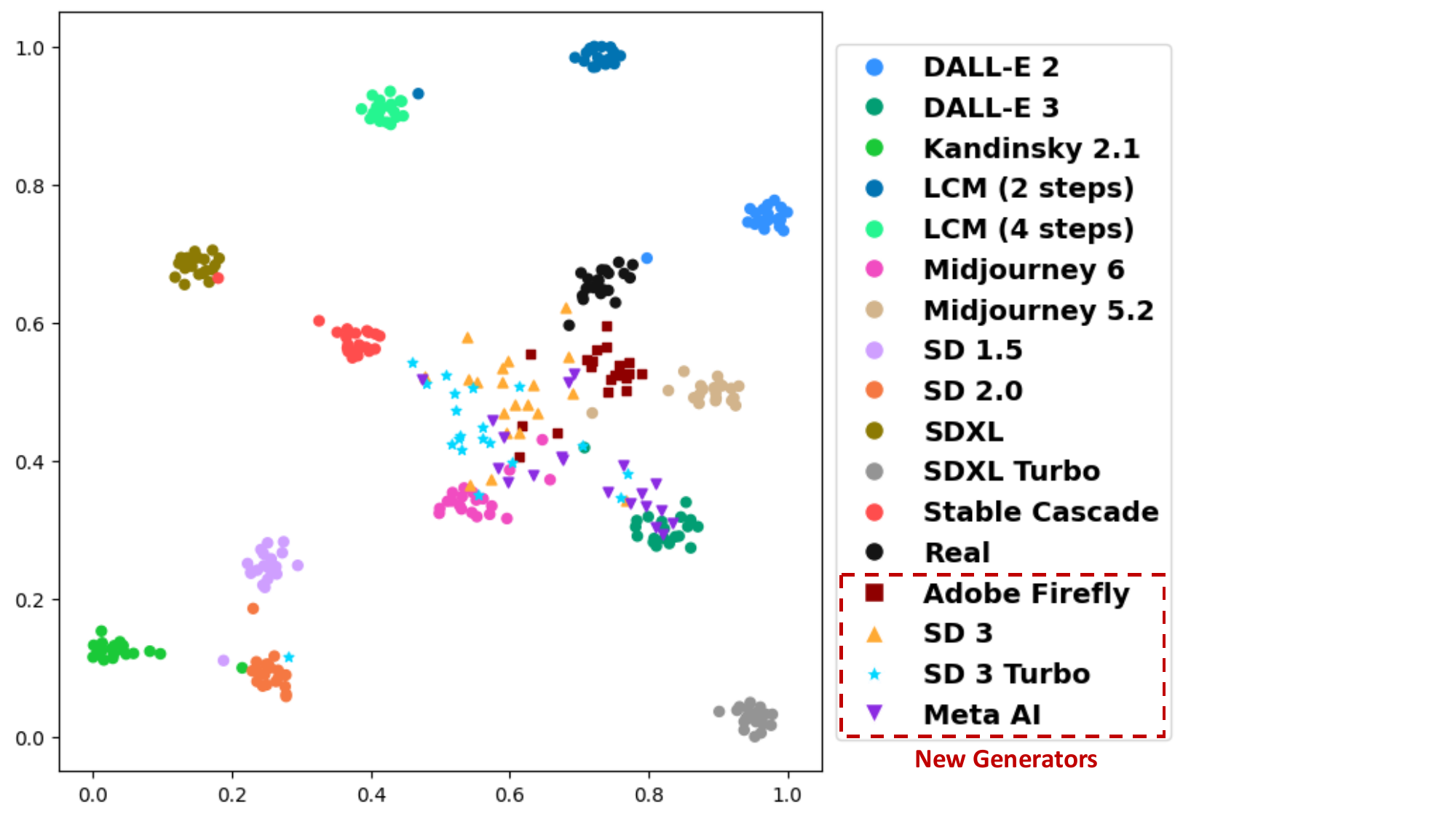}
    \vspace{-20 pt}
    \caption{The t-SNE visualization of 4 unseen new generators in the feature space of our pretrained image attributor.}
    \label{fig:model_stealing}
    \vspace{-15 pt}
\end{figure}

\vspace{-5pt}
It's important to note that our research might facilitate `model stealing,' or the reverse engineering of a model's architecture. As an initial experiment, we projected 20 images generated from each of the four most recent non-open-source models—`Adobe Firefly Image 3' \cite{AdobeFirefly}, `SD 3' \cite{StableDiffusion3}, `SD 3 Turbo' \cite{StableDiffusion3}, and `Meta AI Imagine' \cite{MetaAI}—into the t-SNE feature embedding space of our pretrained image attributor. As illustrated in Fig. \ref{fig:model_stealing}, we observe that images from `Adobe Firefly Image 3' appear similar to those from `Midjourney 5.2' and real images. Meanwhile, `SD 3' and `SD 3 Turbo' are closer to `Stable Cascade' and `Midjourney 6', and `Meta AI Imagine' largely overlaps with `DALL-E 3'. This comparative analysis could lay the groundwork for inferring the architectures of non-open-source models based on those already known.

\arxiv{\vspace{-5pt}}
\section{Elaboration on Results in the Main Paper}
\vspace{-3pt}

\begin{figure*}[!h]
 \vspace{-5 pt}
    \centering
    \includegraphics[trim=0in 0.3in 9.6in 0in, clip,width=0.68\textwidth]{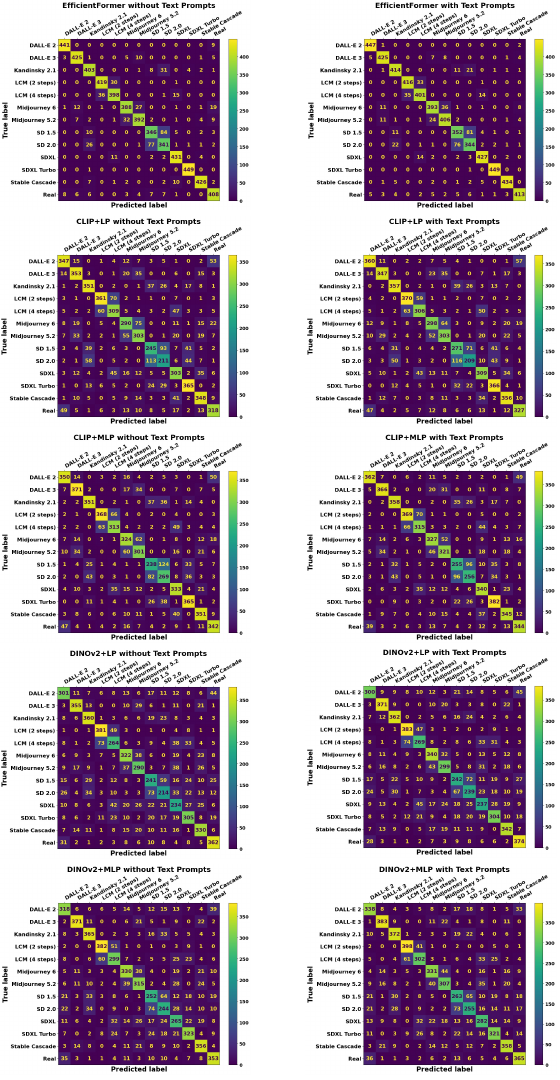}
    \vspace{-10 pt}
    \caption{Confusion matrices for image attributors in Sec. 4.1. The backbone for the CLIP and DINOv2 models is frozen.}
    \label{fig:confusion_image_attributors}
    \vspace{-5 pt}
\end{figure*}

\begin{figure*}[!h]
 \vspace{-5 pt}
    \centering
    \includegraphics[trim=0in 0in 0in 0in, clip,width=\textwidth]{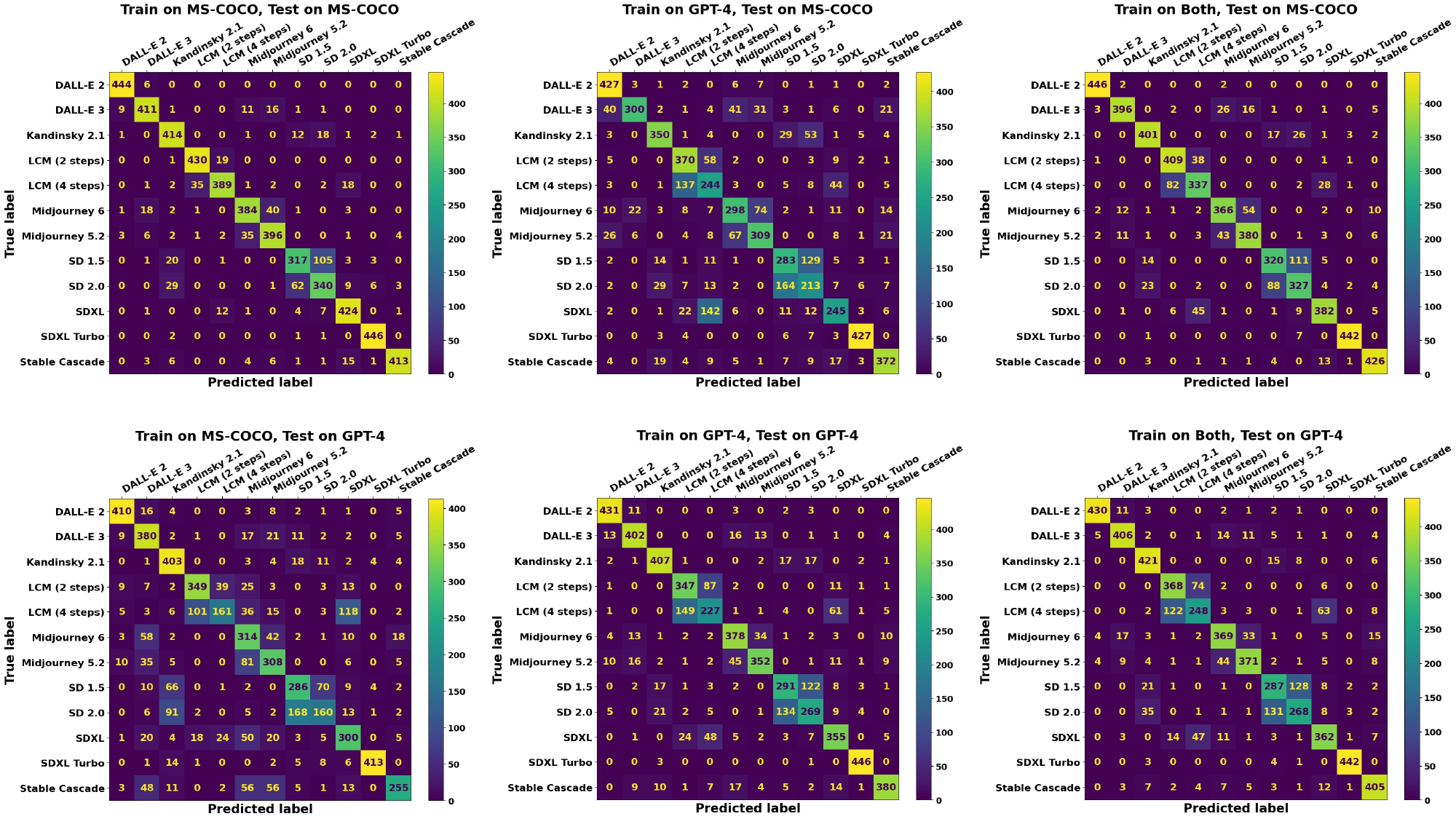}
    \vspace{-15 pt}
    \caption{Confusion matrices for cross-domain generalization in Sec. 4.1.}
    \label{fig:confusion_coco_gpt4}
    \vspace{-5 pt}
\end{figure*}

\begin{figure*}[!h]
 \vspace{-5 pt}
    \centering
    \includegraphics[trim=0in 0in 8.7in 0in, clip,width=0.8\textwidth]{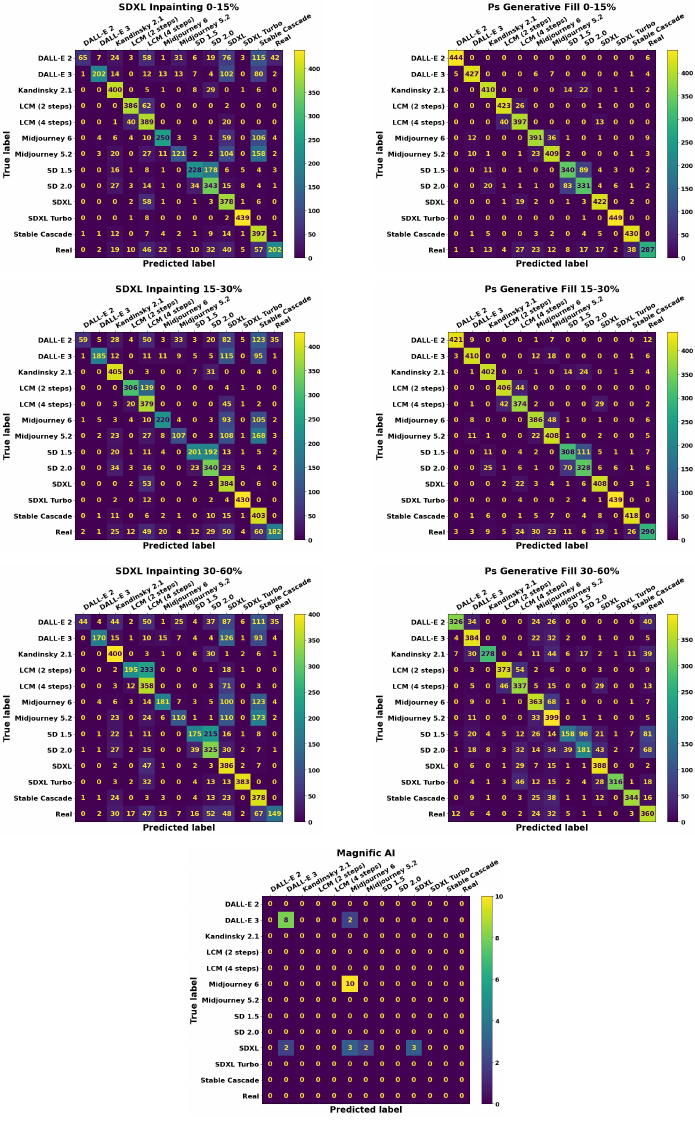}
    \vspace{-12 pt}
    \caption{Confusion matrices for evaluating on post-edited images in Sec. 4.3.}
    \label{fig:confusion_post_editing}
    \vspace{-5 pt}
\end{figure*}

\begin{figure*}[!th]
 \vspace{-5 pt}
    \centering
    \includegraphics[trim=0in 0.8in 0.01in 0in, clip,width=\textwidth]{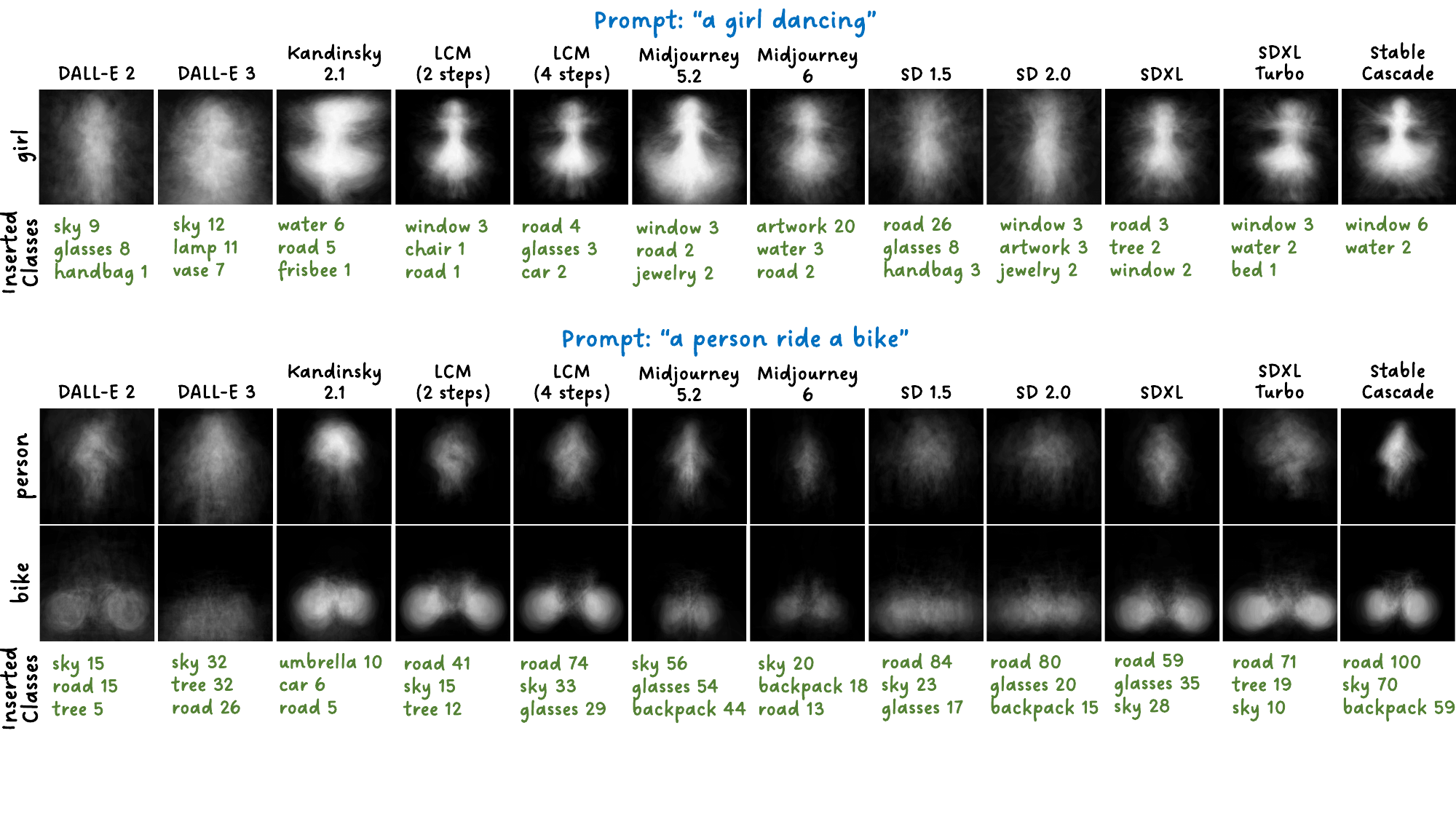}
    \vspace{-18 pt}
    \caption{Additional image composition analyses across generators. We show the averaged segmentation masks for each semantic class indicated on the left side. We also list the top three inserted classes and the number of images (out of 100) with these classes.}
    \label{fig:image_layout_supp}
    \vspace{-5 pt}
\end{figure*}

In this section, we expand upon the results from the experiments performed in the main paper. Figure \ref{fig:confusion_image_attributors} and Table \ref{tab:image_attributor_precision_recall_f1} showcase the confusion matrices and evaluation metrics for the image attributors in Sec. \nonarxiv{4.1}\arxiv{\ref{sec:training_image_attributors}}. Furthermore, Figure \ref{fig:confusion_coco_gpt4} and Table \ref{tab:cross_domain_precision_recall_f1} present the confusion matrices and evaluation metrics for the cross-domain generalization study in Sec. \nonarxiv{4.1}\arxiv{\ref{sec:training_image_attributors}}. Additionally, Figure \ref{fig:confusion_post_editing} illustrates the confusion matrices for post-editing enhancements in Sec. \nonarxiv{4.3}\arxiv{\ref{sec:Assessing_Detectability_Following_Post-Editing_Enhancements}}. Lastly, Figure \ref{fig:image_layout_supp} visualizes the averaged segmentation masks across generators for two additional prompts, which is an extension of our image composition analysis in Sec. \nonarxiv{5}\arxiv{\ref{sec:Detecting_Image_Attribution_Beyond_RGB}}.

\textbf{Takeaways from high-frequency perturbations.}
Prior works have predominantly claimed that classifiers in tasks like `real vs. fake' and image attribution primarily learn from discriminative information in the high-frequency domain. While we concur that high-frequency details can be crucial for discrimination, our work has demonstrated that even when these details are altered, the classifier can still identify highly discriminative features and attain decent accuracy. Our finding does not contradict earlier claims, but rather suggests a shift in perspective, showing that reliance only on high-frequency details may not be necessary.

\begin{table*}[h!]
\vspace{-8 pt}
\centering
\begin{tabularx}{\textwidth}{l| *{4}{>{\centering\arraybackslash}X|} >{\centering\arraybackslash}X}
\toprule
           &  E.F. (scratch) & CLIP+LP     & CLIP+MLP      & DINOv2+LP     & DINOv2+MLP     \\
\midrule
Accuracy   &  90.03/90.96  &  70.15/71.44  &  73.09/74.19  &  67.68/69.44  &  71.33/73.08   \\
Precision  &  90.07/90.98  &  69.95/71.30  &  73.13/74.12  &  67.36/69.09  &  71.20/72.91   \\
Recall     &  90.03/90.96  &  70.15/71.44  &  73.09/74.19  &  67.68/69.44  &  71.33/73.08   \\
F1         &  90.04/90.96  &  70.00/71.25  &  73.07/74.12  &  67.45/69.17  &  71.23/72.93   \\
\bottomrule
\end{tabularx}
\vspace{-8pt}
\caption{Additional quantitative evaluation of image attributors for 13-way classification, consisting of 12 generators and a set of real images. The values (percentages) represent training each attributor \emph{Without / With} text prompts.}
\label{tab:image_attributor_precision_recall_f1}
\vspace{-5 pt}
\end{table*}

\begin{table*}[h!]
\centering
\begin{tabularx}{\textwidth}{l| *{2}{>{\centering\arraybackslash}X|} >{\centering\arraybackslash}X}
\toprule
            & Train on MS-COCO & Train on GPT-4 & Train on Both \\
\midrule
Accuracy    & 89.04/69.24  & 71.07/79.35  & 85.78/81.06  \\
Precision   & 89.07/70.38  & 71.81/79.29  & 85.88/80.87  \\
Recall      & 89.04/69.24  & 71.07/79.35  & 85.78/81.06  \\
F1          & 88.99/68.44  & 71.06/79.21  & 85.78/80.86  \\
\bottomrule
\end{tabularx}
\vspace{-6pt}
\caption{Cross-domain generalization in image attributors. The amount of training and testing data was kept consistent across trials, and an equal number of images was sourced from MS-COCO and GPT-4 prompts for the `Train on Both' trial. The values (percentages) represent testing on images from \emph{MS-COCO / GPT-4} prompts.}
\label{tab:cross_domain_precision_recall_f1}
\vspace{-5 pt}
\end{table*}

\arxiv{\vspace{-26pt}}
\section{Grad-CAM Visualizations}
\vspace{-3pt}
\arxiv{\vspace{-4pt}}

\begin{figure*}[!th]
 \vspace{-5 pt}
    \centering
    \includegraphics[trim=0in 1.8in 0.15in 0in, clip,width=\textwidth]{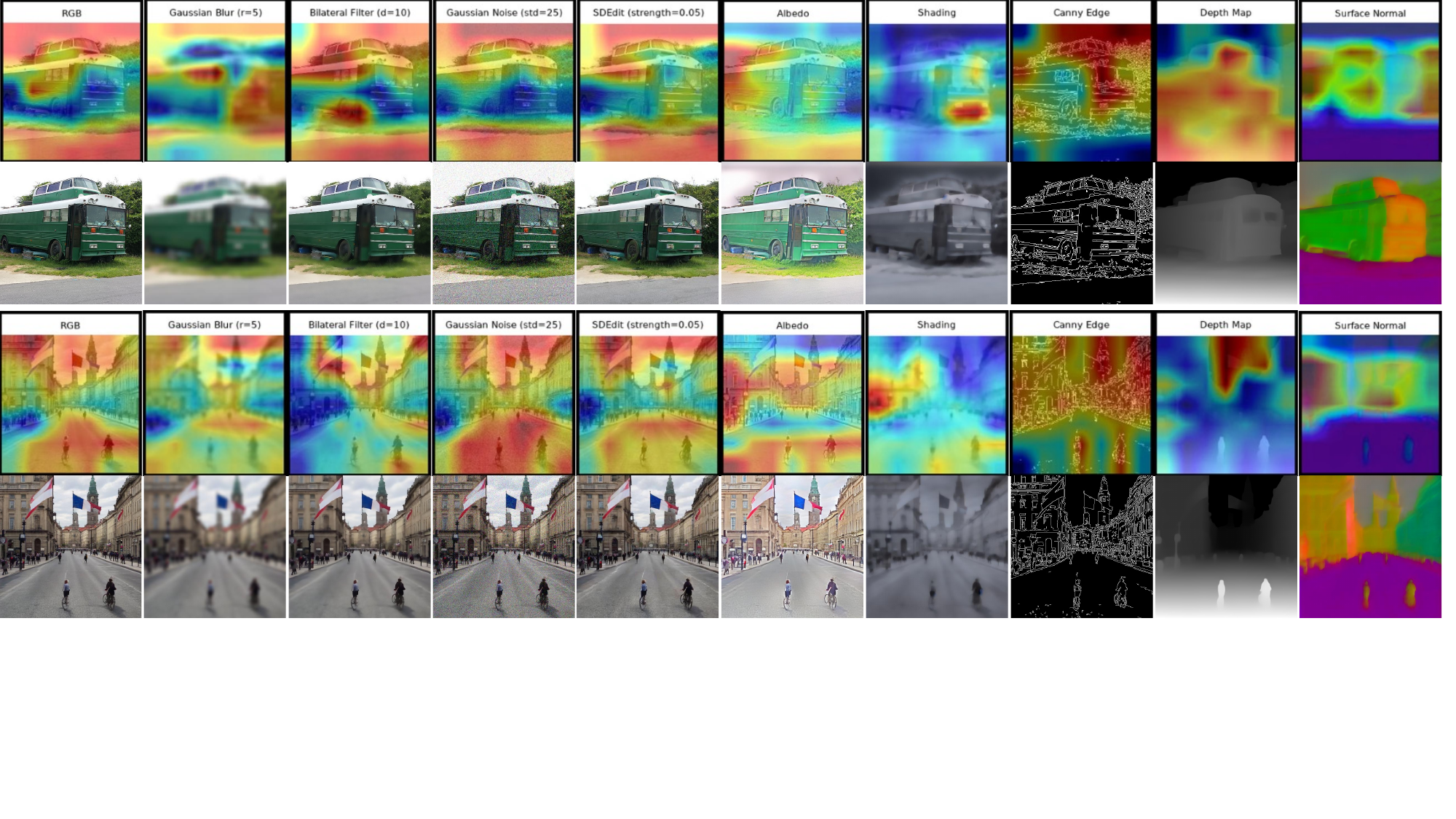}
    \vspace{-15 pt}
    \caption{Grad-CAM \cite{Selvaraju_2019_gradcam, jacobgilpytorchcam} visualizations for image attributors trained on each image type, where each column represents a distinct attributor. The first and third rows illustrate the Grad-CAM heatmaps overlaid on the input images. The second and fourth rows show the input images without Grad-CAM. The first example on the top is based on a real image from MS-COCO \cite{lin2015microsoft_coco_dataset}, while the second example on the bottom is based on a fake image generated by SDXL Turbo \cite{sauer2023adversarial}. We notice that the attributors trained on RGB images and images after high-frequency perturbations often focus on relatively smooth image regions, such as the sky or ground.}
    \label{fig:gradcam_supp}
    \vspace{-5 pt}
\end{figure*}

Figure \ref{fig:gradcam_supp} showcases the Grad-CAM \cite{Selvaraju_2019_gradcam, jacobgilpytorchcam} heatmaps for image attributors trained on various image types, including the original RGB images, images after high-frequency perturbations, and mid-level representations. We observe that the image attributors trained on RGB images and images after high-frequency perturbations tend to pay attention to smooth image regions, such as the sky or ground. Nonetheless, even though the attributors focus on varied image regions, it remains difficult to explain how they make their decisions for each image.

\section{Broader Impacts}
We acknowledge that text-to-image diffusion models pretrained on large-scale, uncurated web data may produce biases and errors. Additionally, we use text prompts that are based on captions of MS-COCO \cite{lin2014microsoft} images and GPT-4 \cite{achiam2023gpt} outputs, which may generate images of people.

\end{document}